\documentclass{article}

\PassOptionsToPackage{numbers, sort&compress}{natbib}
\usepackage[nonatbib, preprint]{neurips_2023}

\usepackage[utf8]{inputenc} 
\usepackage[T1]{fontenc}    
\usepackage{hyperref}       
\usepackage{url}            
\usepackage{booktabs}       
\usepackage{amsfonts}       
\usepackage{nicefrac}       
\usepackage{microtype}      
\usepackage{xcolor}         
\usepackage{comment}
\usepackage{bm}
\usepackage{enumitem}
\usepackage{amsmath}        
\usepackage{mathtools}
\usepackage{wrapfig, lipsum}
\usepackage{adjustbox}

\usepackage{graphicx}
\usepackage{multirow, booktabs}
\usepackage[small, bf]{caption}
\usepackage{subcaption}
\usepackage{makecell}

\DeclarePairedDelimiterX{\infdivx}[2]{\{}{\}}{%
	#1\;\delimsize\|\;#2%
}

\usepackage[
    style=numeric, 
    maxbibnames=99,
    sorting=nyt,
]{biblatex}
\title{Who Needs Decoders? Efficient Estimation of Sequence-level Attributes}

\author{%
  Yassir Fathullah\\
  Engineering Department\\
  University of Cambridge\\
  \texttt{yf286@cam.ac.uk} \\
  \And
  Adian Liusie\\
  Engineering Department\\
  University of Cambridge\\
  \texttt{al826@cam.ac.uk} \\
  \AND
  Puria Radmard\\
  Engineering Department\\
  University of Cambridge\\
  \texttt{pr450@cam.ac.uk} \\
  \And
  Mark J. F. Gales\\
  Engineering Department\\
  University of Cambridge\\
  \texttt{mjfg@eng.cam.ac.uk} \\
}

\author{%
  Yassir Fathullah, Puria Radmard, Adian Liusie, Mark J. F. Gales\\
  Engineering Department, University of Cambridge\\
  \texttt{\{yf286, pr450, al826, mjfg100\}@cam.ac.uk} \\
}

\makeatletter
\def\ps@myheadings{%
    \let\@oddfoot\@empty\let\@evenfoot\@empty
    \def\@evenhead{\thepage\hfil\slshape\leftmark}%
    \def\@oddhead{{\slshape\rightmark}\hfil\thepage}%
    \let\@mkboth\@gobbletwo
    \let\sectionmark\@gobble
    \let\subsectionmark\@gobble
    }
  \if@titlepage
  \renewcommand\maketitle{\begin{titlepage}%
  \let\footnotesize\small
  \let\footnoterule\relax
  \let \footnote \thanks
  \null\vfil
  \vskip 60\p@
  \begin{center}%
    {\LARGE \@title \par}%
    \vskip 3em%
    {\large
     \lineskip .75em%
      \begin{tabular}[t]{c}%
        \@author
      \end{tabular}\par}%
      \vskip 1.5em%
    {\large \@date \par}
  \end{center}\par
  \@thanks
  \vfil\null
  \end{titlepage}%
  \setcounter{footnote}{0}%
}
\else
\renewcommand\maketitle{\par
  \begingroup
    \renewcommand\thefootnote{\@fnsymbol\c@footnote}%
    \def\@makefnmark{\rlap{\@textsuperscript{\normalfont\@thefnmark}}}%
    \long\def\@makefntext##1{\parindent 1em\noindent
            \hb@xt@1.8em{%
                \hss\@textsuperscript{\normalfont\@thefnmark}}##1}%
    \if@twocolumn
      \ifnum \col@number=\@ne
        \@maketitle
      \else
        \twocolumn[\@maketitle]%
      \fi
    \else
      \newpage
      \global\@topnum\z@   
      \@maketitle
    \fi
    \thispagestyle{plain}\@thanks
  \endgroup
  \setcounter{footnote}{0}%
}
\makeatother

\begin{document}

\maketitle

\begin{abstract}
    State-of-the-art sequence-to-sequence models often require autoregressive decoding, which can be highly expensive. 
    %
    However, for some downstream tasks such as out-of-distribution (OOD) detection and resource allocation, the actual decoding output is not needed just
    a scalar attribute of this sequence.
    In these scenarios, where for example knowing the quality of a system's output to predict poor performance prevails over knowing the output itself, is it possible to bypass the autoregressive decoding?
    We propose Non-Autoregressive Proxy (NAP) models that can efficiently predict general scalar-valued sequence-level attributes.
    Importantly, NAPs predict these metrics directly from the encodings, avoiding the expensive autoregressive decoding stage.
    We consider two sequence-to-sequence task: Machine Translation (MT); and Automatic Speech Recognition (ASR).
    In OOD for MT, NAPs outperform a deep ensemble while being significantly faster.
    NAPs are also shown to be able to predict performance metrics such as BERTScore (MT) or word error rate (ASR). 
    For downstream tasks, such as data filtering and resource optimization, NAPs generate performance predictions that outperform predictive uncertainty while being highly inference efficient.
\end{abstract}

\section{Introduction}
Autoregressive encoder-decoder models have emerged as the dominant approach for many sequence-to-sequence tasks \cite{sutskever2014sequence} and are the state-of-the-art for a range of tasks such as Automatic Speech Recognition (ASR) \cite{conformer, whisper}, Machine Translation (MT) \cite{transformers, mt5}, and Abstractive Text Summarization \cite{flan-palm, t5}. 
However, for many applications, the decoded output sequence is not actually required, only attributes of the sequence. 
In out-of-distribution (OOD) detection, only a sequence-level metric such as confidence is required \cite{hendrycksbaseline, structured}.
In selective classification \cite{deepselective, xiaselective, selective} the output is only needed if the prediction is trusted and rejected in all other cases. 
Another example is deferral strategies for resource allocation \cite{cnn-cascade, early-exit, cascade, xiawindow-cascade, cascade-v2}, where computation is allocated between systems of different complexity. Standard deferral strategy approaches use the predictive uncertainty of a simpler system to decide whether or not to pass it on to a better-performing system of higher complexity \cite{wang2022wisdom}. 
%


All of the examples above require some form of predictive uncertainty metric from the output, which in the case of autoregressive models are expensive to obtain \cite{wu2016google}.
Transformers, for instance, have recently dominated many sequence-to-sequence tasks \cite{gpt3, palm, t5}, but are often extremely large, with several million to billions of parameters.
Combined with the quadratic cost of self-attention \cite{transformers} and autoregressive decoding (equipped with beam-search \cite{koehn}), this  can limit the application of these systems in real-world settings, such as those that have limited computational resources or require low latency \cite{cascade}. 
Furthermore, ensembling generally improves system performance and can be leveraged for useful analysis, such as for robust uncertainty estimation \cite{gal2016dropout, deepensembles}. 
However, ensembles' memory and inference costs scale linearly with the number of members in the ensemble, making them even more impractical for real-world scenarios.
There are methods including Knowledge Distillation (KD) \cite{seq-training, kd} and Ensemble Distribution Distillation (EDD) \cite{edd, edd-gec} that attempt to distil the knowledge from an autoregressive ensemble into a single neural network but this still does not circumvent the high costs associated with autoregressive generation.

%
%
%
%

Previous works have investigated adding a second output head explicitly trained to capture a specific metric such as epistemic uncertainty in image segmentation \cite{dudes} or the true class probability in image classification \cite{truecp}. 
The work of \cite{cem} extends this style of approach to ASR by adding a second head to the decoder, to predict token-level decoding errors. Despite its success in providing robust estimates, computing the output uncertainties still requires an expensive autoregressive decoding process. 
The work of \cite{al-proxy} trains an independent proxy model for estimating uncertainties. This method is based on training a much smaller image classification model in an identical manner to the primary model, instead using the uncertainty estimates produced by the small model's outputs to guide the primary one. In the space of autoregressive encoder-decoder models, this approach is still not feasible; the costs of training and decoding persist even for small autoregressive models.

In this paper we propose \textit{Non-Autoregressive Proxy} (NAP) models that directly estimate sequence-level attributes, bypassing the expensive decoding process of autoregressive systems.
%
%
When deployed, these lightweight proxy models can be used to robustly predict sequence properties using a fraction of the computational requirements.
Our approach is kept general and applicable to any sequence attribute, demonstrating the usefulness of this framework to diverse metrics such as sequence-level predictive uncertainty, BERTScore for MT and word error rate (WER) for ASR.
Investigations into downstream tasks such as out-of-distribution (OOD) detection show that NAPs can outperform an ensemble at fraction of the inference time. Due to the flexibility of the proposed framework, we also investigate training NAPs on sequence-level performance metrics (BERTScores and WERs), outperforming uncertainty-based approaches to data filtering and resource optimization.



%
%
%
%

\section{Background}


For sequence modelling tasks, there has been a range of work on predicting sequence-level attributes. One common example is deriving information theoretic uncertainties from the outputs of autoregressive systems \cite{structured, notin2021improving}, where unsupervised token-level uncertainties from some decoding process are combined to form sequence-level estimates. Such sequence-level uncertainties are then used in downstream tasks such as OOD detection \cite{structured}, quality estimation \cite{fomicheva2020unsupervised} and curriculum learning \cite{zhou2020uncertainty}.

Previous work has also explored task-specific supervised approaches to confidence/metric estimation. The work of \cite[t]{gamper2020predicting} explores training a small independent model to predict the sub-utterance-level word error rate (WER) of a primary ASR model for short-duration audio when the reverberant conditions change. However, the approach is not generalizable to other domains such as MT due to the specific focus on reverberant speech.
Other work has also focused on training an error detection module attached to the decoder of some ASR or MT system \cite{confidence-scores-asr, koehn, nmt-calibration, cem, asr-confidence-decoding, asr-confidence-bilstm}. For example, a typical approach to training the decoder-side error detector is based on token-level error labels from the minimum Levenshtein distance alignment to the ground truth. From these token-level estimates, a sequence-level confidence score can be derived. In ASR where there is often one clear true transcription of the input audio, such an error detection module is appropriate. However, these approaches are inappropriate for MT where multiple translations could all have the same meaning and be considered valid. Such a token-level error detector would flag other valid translations as errorful even when conveying the same information and meaning.

This final example is one of the main motivations behind BERTScore and related approaches \cite{bleurt, bartscore, bertscore, zhao2019moverscore}. BLEU \cite{papineni2002bleu, sacrebleu} has long been the main MT evaluation metric for measuring sequence similarity between a translation and a reference using some measure of overlap. However, it suffers from similar issues as (Levenshtein) edit-distance metrics. BERTScore resolves such issues by leveraging bidirectional language models in generating contextual variable-length embeddings for both the translation and reference sequence, computing an automatic sequence similarity score in this embedding space. There has also been a set of work on supervised MT quality estimation \cite{specia-etal-2020-findings-wmt, specia-etal-2021-findings, zerva-etal-2022-findings} in which models are trained to estimate the quality (human expert estimated metric) of a translation by making use of the source, the decoded translation and additional token-level probability. However, both the automatic BERTScore and quality metrics require an expensive autoregressive decoding stage to obtain the estimate.

\section{Non-Autoregressive Proxy (NAP): Efficient Sequence Metric Estimation}
We are interested in the general problem of estimating sequence-level attributes whilst remaining highly inference-efficient. 
These sequence-level metrics include: (1) information-theoretic uncertainties \cite{structured}; (2) neural-based evaluation scores such as BERTScore \cite{bertscore}; and (3) discrete sequence-similarity metrics such as word error rate. The standard approach to obtaining these sequence-level metrics is to run an expensive autoregressive decoding scheme to produce a set of hypotheses. One can either extract sequence attributes directly from this hypothesis set \cite{structured} or compare them with their corresponding references to obtain a measure of sequence similarity.

\begin{figure}[h!]
     \centering
     \begin{subfigure}[b]{0.4721\textwidth}
         \centering
         \includegraphics[width=\textwidth]{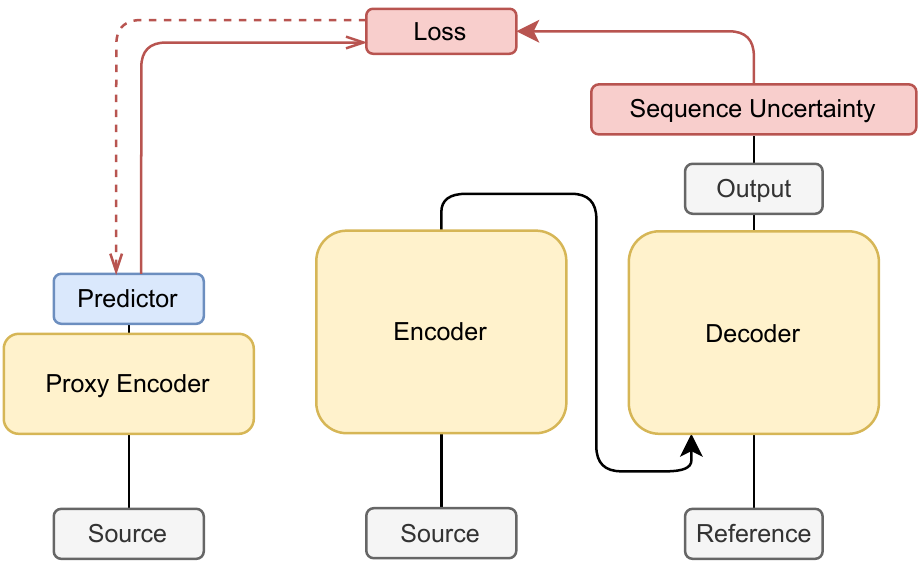}
         \caption{Setup 1: Capturing sequence uncertainties.}
         \label{fig:proxy-setup-v1}
     \end{subfigure}
     \hfill
     \begin{subfigure}[b]{0.5186\textwidth}
         \centering
         \includegraphics[width=\textwidth]{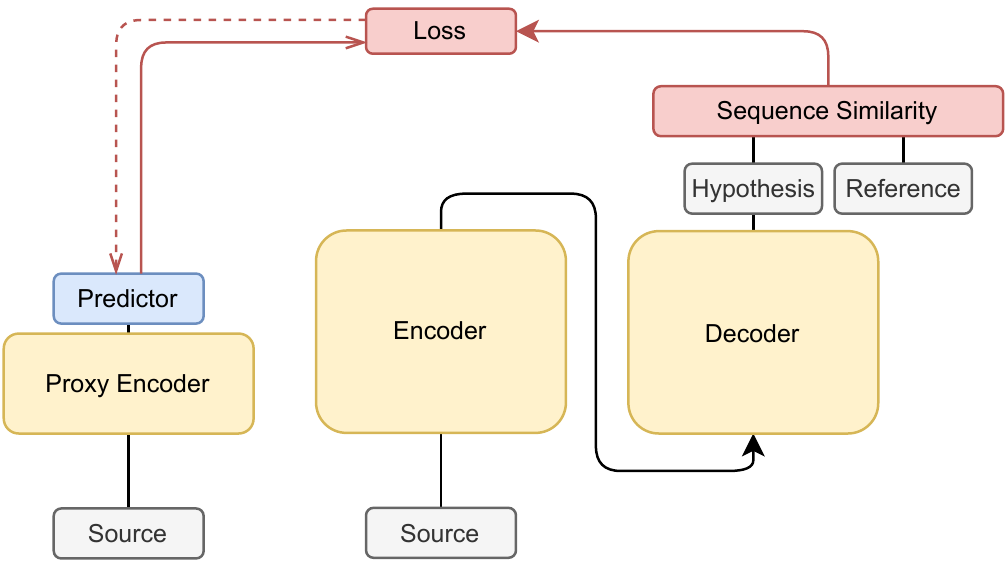}
         \caption{Setup 2: Capturing sequence similarities.}
         \label{fig:proxy-setup-v2}
     \end{subfigure}
    \caption{Our proposed proxy training scheme: A teacher encoder-decoder model trains a proxy encoder student to predict consistent sequence scores using some loss function. In \textbf{(a)} we train the proxy to extract sequence uncertainties from a decoder that is fed the reference. In \textbf{(b)} we train a proxy to capture sequence-level similarity scores (e.g. BERTScore or WER) from decoded outputs.}
    \label{fig:proxy-setup}
\end{figure}

The aim of this paper is to avoid the costly autoregressive generation stage and instead train an encoder-only, non-autoregressive proxy (NAP) model to directly imitate the sequence metrics produced by an autoregressive system, using only the source, see Figure \ref{fig:proxy-setup}. 

We employ two different setups as shown in Figures \ref{fig:proxy-setup-v1} and \ref{fig:proxy-setup-v2}. The aim of the first setup is to train a proxy to directly extract sequence uncertainties when the main model is additionally given the reference sequence. This is in order to teach the proxy model to imitate the uncertainties from the gold reference. The second setup aims to teach the proxy a sequence similarity score when the autoregressive generated hypothesis is compared to the reference. Both setups are highly challenging as the non-autoregressive proxy is tasked with predicting sequence-level metrics from only the source. 


However, the key feature of the NAP is that it directly predicts these metrics without a decoding scheme (e.g. beam search) and without any reference sequences, allowing the user to extract useful information from large amounts of unlabelled data with little cost. Furthermore, in the first setup of Figure \ref{fig:proxy-setup-v1}, the proxy also avoids the exposure bias problem \cite{scheduled-sampling, seq-training}, by directly training on the teacher-forced \cite{teacher-forcing} sequence uncertainties. 

%
In this work, we follow Figure \ref{fig:proxy-setup-v1} in training a proxy on both single teacher confidence and entropy scores or ensemble mutual information estimates, evaluating its imitation ability and downstream out-of-distribution detection ability. We also follow Figure \ref{fig:proxy-setup-v2} in training a proxy to predict BERTScores in Machine Translation and WER in Speech Recognition and evaluate the performance of the NAP on a data filtering and resource optimization task.

\textbf{Loss Function:} Sequence-level metrics are represented by single scalar values. Therefore, the proxy student can be trained using any regression loss function. However, unlike standard regression tasks, we seek to learn the relative ordering (rankings) of our scores, as this simplifies the task and is more pertinent for downstream applications such as OOD detection. Therefore, we will mainly opt for the Spearman Rank and Pearson correlation coefficient (SCC \& PCC) depending on the specific task considered. Consider a batch of $n$ items with teacher scores $\{ s_i \}_{i = 1}^{n}$ and corresponding proxy predictions $\{ \hat{s}_i \}_{i = 1}^{n}$. The loss functions are then defined as:
\begin{equation}
    \mathcal{L}_{\tt SCC} = - \left(1 - \frac{6\sum_{i}(r(s_i) - r(\hat{s}_i))^2}{n(n^2 - 1)} \right) \hspace{5mm} \mathcal{L}_{\tt PCC} = - \left( \frac{\sum_i (s_i - \mu_{s})(\hat{s}_i - \mu_{\hat{s}})}{\sqrt{\sum_i (s_i - \mu_{s})^2}\sqrt{\sum_i (\hat{s}_i - \mu_{\hat{s}})^2}} \right)
\end{equation}
where $r(s) \in \{1, 2, \dots, n\}$ signifies the rank of $s$. Since the rank operator is discrete and non-differentiable it is not directly applicable to our application. We resort to a differentiable Spearman Rank extension \cite{sorting} with an open source implementation\footnote{\url{https://github.com/google-research/fast-soft-sort}}. The Pearson loss can directly be applied without any modifications. 
Note that unlike its original usage \cite{sorting}, where the system is trained to rank class values for a single instance, we are using this loss to sort single values associated with multiple different items in a batch. 
We also investigate alternative loss functions such as the root mean squared error (RMSE) and mean absolute error (MAE), see Appendix B.1.

\textbf{Predictor Design:} In order to produce a scalar score from a variable-length encoder-output representation, we make use of a pooling operation. We utilize two options, temporal averaging or multi-head attention with a single trainable query. The encoder vector outputs $\{{\bm v}_l\}_{l = 1}^{L}$ are therefore pooled to form a fixed-size representation ${\bm v}$ which is fed into a three-layer multi-layer perception (MLP). Furthermore, early exploratory experiments found that a softmax activation is vital for good performance as it can be seen as introducing inductive bias into the estimation of information-theoretic and related metrics. Architectural details of the MLP and ablation studies are provided in Appendix B.2.
%

\textbf{Proxy Encoder Backbone:} By default, the NAP encoder backbone is initialized from the encoder weights of the main encoder-decoder model. However, many pretrained models such as the T5 \cite{t5} and Whisper \cite{whisper} come in different sizes. The backbone will therefore also be initialized from the encoder of a smaller T5 model and be taught to predict the uncertainties from a larger system. Appendix B.4 further explores `mismatched' encoders, e.g. using a RoBERTa NAP to predict the output attributes of a T5 system.

Furthermore, all experiments in this paper freeze the encoder backbone and only train the small predictor on top of the NAP encoder. This improves the training speed and memory usage allowing a user to train multiple predictor heads on top of the same backbone, each for a different metric (e.g. estimating sequence-level confidence and BERTScores in the same forward pass). Note that the purpose of our investigations are not to create the best possible NAP model (for example, fine-tuning the backbone encoder could improve performance at no cost of inference speed). We only seek to demonstrate that this approach is highly flexible and applicable to a range of sequence-level metrics and can provide cheap but useful information for sequence-to-sequence tasks.

\section{Experimental Evaluation}

\textbf{Predicting Uncertainties}: We will evaluate the imitation ability of NAP models on various tasks. Following Setup 1, the first set of experiments will focus on the ability of a proxy system to capture sequence-level confidence or entropy from a single T5 transformer \cite{t5} finetuned on a spoken-language Machine Translation (MT) dataset. We further explore the ability of NAPs to imitate mutual information (epistemic uncertainty \cite{der2009aleatory, hora1996aleatory}) from an ensemble of T5 systems. The performance of the NAPs will then be evaluated by measuring the Spearman Rank correlation between the teacher (under teacher-forcing \cite{teacher-forcing}) and the proxy estimates on a range of in-domain (ID) and out-of-domain (OOD) datasets. We also investigate the performance of the proposed NAP on OOD detection.

\textbf{Predicting BERTScores}: Following Setup 2, we also investigate if proxy systems can capture much more complex sequence metrics such as BERTScores \cite{bertscore} from a single T5 in MT. Capturing this metric is especially challenging since the beam-search output of the T5 decoder and corresponding reference will be fed through a language model such as BERT \cite{bert} which then computes the final score. The performance will be measured by computing the Spearman Rank between proxy outputs and BERTScores on both ID and OOD datasets. Furthermore, the proxy is compared to sequence-level confidence and entropy scores from the T5 model to see how well they correlate with BERTScores. 

\begin{figure}[h!]
    \centering
    \begin{subfigure}[b]{0.341\textwidth}
        \centering
        \includegraphics[width=\textwidth]{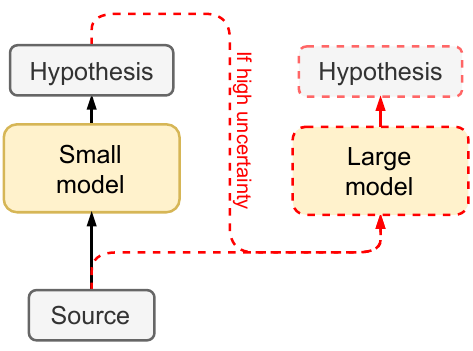}
        \caption{Baseline deferral system.}
        \label{fig:deferral-baseline}
    \end{subfigure}
    \hfill
    \begin{subfigure}[b]{0.550\textwidth}
        \centering
        \includegraphics[width=\textwidth]{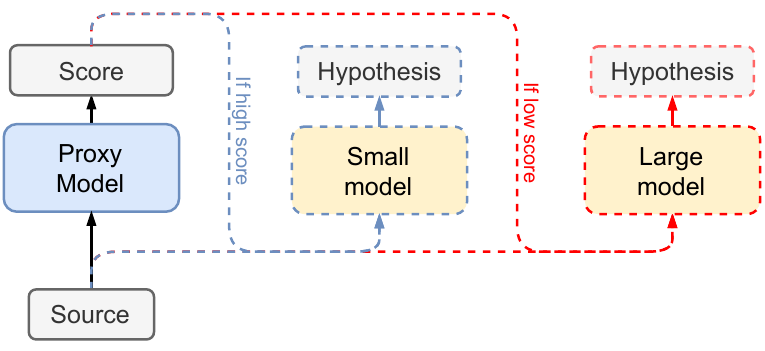}
        \caption{Proxy deferral system.}
        \label{fig:deferral-proxy}
    \end{subfigure}
    \caption{Deferral systems: In the baseline deferral system, the inputs with high uncertainty (under the small model) are fed into the larger model. In the proxy deferral system, model selection is based on the output of the efficient proxy.}
    \label{fig:deferral-systems}
\end{figure}

The performance of a BERTScore estimating proxy system can also be evaluated on two downstream tasks: \textit{Filtering task} \cite{cem}: Given a dataset, we remove the examples with the lowest proxy or highest uncertainty estimate. For good estimates, the filtered subset should display a higher average BERTScore. \textit{Resource optimization task} \cite{cascade}: Under a fixed resource budget, one seeks to allocate inputs to models of different complexity in order to maximise performance. A well-performing allocation system would achieve higher performance with a smaller budget, see Figure \ref{fig:deferral-systems}.

\textbf{Predicting WER}: Finally, we follow Setup 2 in investigating if a NAP can imitate the sentence-level WER and the total number of errors produced by an ASR system. In this case, we utilize the pretrained state-of-the-art Whisper \cite{whisper} models on the LibriSpeech corpus \cite{librispeech}. Since the Whisper model is very well-performing, it is able to perfectly decode a large fraction of the dataset, which would cause issues for a rank-based loss such as Spearman. We, therefore, resort to Pearson for these experiments. Note, the corpus-level WER performance of an ASR system is a length-weighted average of the sentence-level WERs. Therefore, we also train NAPs to predict the number of decoding errors in an utterance. Similar to the BERTScore experiments, the performance of NAPs will be evaluated in a similar manner using both filtering and resource optimization tasks.

\subsection{Machine Translation}

We use the IWSLT 2017 English-to-German training set for finetuning T5 systems on spoken language translation. We generate a three-model ensemble of T5 systems which we use as a stronger baseline for uncertainty estimation. We also investigate if Knowledge Distillation (KD) \cite{kd} and Ensemble Distribution Distillation (EDD) \cite{edd, edd2} are able to imitate the uncertainties produced by a single or ensemble systems respectively.

We use a range of in-domain and out-of-domain datasets for downstream tasks. These include the Web Inventory Talk (Ted IWSLT 2016; ID), Newstest19 \& 20 news commentary (OOD-1), Khresmoi medical data (OOD-2), MTNT-2019 Reddit text (OOD-3) and KFTT Kyoto-related Wikipedia articles (OOD-3) datasets. All but the latter two datasets are English-to-German, while the final two are English-to-Japanese. Due to the language mismatch, OOD-3 datasets cannot be used to evaluate BERTScore prediction in Section \ref{sssec:bertscores-mt}. Dataset and setup details are available in Appendix A.

Table \ref{tab:speed} shows how long inference of the {\tt iwslt-2017} test set takes for various models. 
This demonstrates a primary desideratum of a NAP, the ability to quickly process large amounts of data.
For example, a large proxy being 46x faster than a T5 Large model using a beam of $B = 12$ (used in experiments below) and is approximately 138x faster than the three-model ensemble (if run serially).
Given the shared architecture between the proxy and primary model encoders, this vast difference in inference time is due to the ability to bypass expensive decoding. 

\begin{table}[h!]
	\centering{}
	\begin{minipage}[t]{1.0\textwidth}%
		\begin{center}
                \caption{Approximate corpus inference time for {\tt iwslt-2017} using Hugging Face \cite{huggingface}, with an {\tt NVIDIA A100}. The proxy performs an encoder forward pass and the T5 models perform beam search. {\tt BERTScore} performance computed for the {\tt B = 12} setting which is used in downstream tasks.}
                \vspace{1mm}
                \small
    		\begin{tabular}{r|ccccc|c}
    			\toprule
                    \multirow{2}{*}{{\tt Model}} & \multicolumn{5}{c|}{{\tt T5 Model}} & \multirow{2}{*}{\tt NAP} \\
                    & {\tt B = 1} & {\tt B = 4} & {\tt B = 8} & {\tt B = 12} & {\tt BERTScore} \\
                    \midrule
                    {\tt Small (S)} & 41.9s & 85.9s & 119.7s & 178.6s & 67.4 & 2.7s \\
                    {\tt Base (B)} & 117.7s & 270.3s & 347.4s & 537.6s & 68.2 & 5.5s \\
                    {\tt Large (L)} & 313.7s & 583.4s & 755.0s & 826.6s & 68.6 & 17.9s \\
                    %
    			\bottomrule
    		\end{tabular}
			\label{tab:speed}
		\end{center}
	\end{minipage}
\end{table} 

\subsubsection{Estimating Uncertainties in Machine Translation}


We trained NAPs (of three different sizes, see Table \ref{tab:speed}) to predict either sequence-level confidence $\mathcal{P}$ or entropy $\mathcal{H}$ (using the conditional approximation described in \cite{structured}) of a T5 Large model. We also trained NAPs to predict the mutual information $\mathcal{I}$ score produced by an ensemble of finetuned T5 Large models. The performance of the proxies is compared to two baseline systems: KD when capturing confidence or entropy of a single model, and EDD in capturing mutual information from an ensemble. 
The autoregressive distilled baselines will also be of three different sizes, see Table \ref{tab:speed}.


%
In the case of confidence $\mathcal P$ and mutual information scores $\mathcal I$, the proxy achieves a better rank ordering of instances for both datasets and at all sizes than the corresponding encoder-decoder student, despite being an order of magnitude faster at inference (Table \ref{tab:spearman-baseline}). Knowledge-distilled models are better at imitating their teacher's $\mathcal H$, however, this is not indicative of downstream task performance such as OOD detection, as explored below (Table \ref{tab:detection-performance}). Note that the NAP here is unique in its ability to predict any scalar sequence metric, whereas KD is unable to mimic mutual information scores.

\begin{table}[h!]
	\centering{}
	\begin{minipage}[t]{1.0\textwidth}%
		\begin{center}
                \caption{Spearman Rank correlation of uncertainties when comparing baseline distillation and proxy to the teacher ensemble. Averaged over 3 runs. Standard deviations in the order of $\pm 1.0$.}
			\vspace{1mm}
                \small
    		\begin{tabular}{c|ccc|ccc||ccc}
    			\toprule
                    \multirow{1}{*}{\textbf{Model Size}} & {\tt S} & {\tt B} & {\tt L} & {\tt S} & {\tt B} & {\tt L} & {\tt S} & {\tt B} & {\tt L} \\ 
                    \midrule
                    \textbf{Dataset} & \multicolumn{3}{c|}{\textbf{Distillation $\mathcal{P}$}} & \multicolumn{3}{c||}{\textbf{Distillation $\mathcal{H}$}} & \multicolumn{3}{c}{\textbf{EDD $\mathcal{I}$}} \\
                    {\tt iwslt-2017} & 
                    18.7 & 19.8 & 20.8 & 69.4 & 73.1 & 74.5 & 43.7 & 51.5 & 55.1 \\
                    {\tt ted-iwslt-2016} & 
                    21.4 & 21.1 & 21.8 & 57.5 & 59.5 & 60.6 & 46.8 & 47.0 & 48.0 \\
                    \midrule
                    \textbf{Dataset} & \multicolumn{3}{c|}{\textbf{NAP $\mathcal{P}$}} & \multicolumn{3}{c||}{\textbf{NAP $\mathcal{H}$}} & \multicolumn{3}{c}{\textbf{NAP $\mathcal{I}$}} \\
                    {\tt iwslt-2017} & 
                    39.9 & 42.6 & 42.1 & 40.4 & 58.8 & 62.7 & 53.7 & 54.3 & 55.6\\
                    {\tt ted-iwslt-2016} & 
                    26.2 & 25.3 & 25.2 & 44.8 & 52.3 & 53.8 & 50.0 & 49.7 & 51.3 \\
    			\bottomrule
    		\end{tabular}
			\label{tab:spearman-baseline}
		\end{center}
	\end{minipage}
\end{table} 

%
%
%
Finally, we perform downstream out-of-distribution detection using confidence, entropy, and MI scores from T5 Large ensemble, EDD (T5 Large), and Proxy Large. We use {\tt iwslt-2017} as in-domain and measure performance with {\tt AUROC} (a score of 50\% represents random detection). 
Results in Table \ref{tab:detection-performance} show that in all but one scenario, the uncertainty scores predicted by the proxy model are best suited for the task, particularly considering inference speeds. Note that overall, the detection performance of a NAP exceeds that of the Deep Ensemble. A potential explanation is that the proxy is directly trained to predict uncertainties while the ensemble estimates uncertainties based on the beam-search decoded outputs \cite{structured}, suffering from exposure bias \cite{scheduled-sampling, seq-training}.

%
%
%
%

\begin{table}[h!]
	\centering{}
	\begin{minipage}[t]{1.0\textwidth}%
		\begin{center}
                \caption{\%{\tt AUROC} detection performance of autoregressive and proxy models using various uncertainties. Averaged over 3 runs. Standard deviations in the order of $\pm 2.0$.}
			\vspace{1mm}
                \small
    		\begin{tabular}{ll|ccc|ccc|ccc}
    			\toprule
                    \multirow{2}{*}{\textbf{Split}} & \multirow{2}{*}{\textbf{Dataset}} & \multicolumn{3}{c|}{\textbf{Deep Ensemble}} & \multicolumn{3}{c|}{\textbf{EDD}} & \multicolumn{3}{c}{\textbf{NAP}} \\
                    & & $\mathcal{P}$ & $\mathcal{H}$ & $\mathcal{I}$ 
                    & $\mathcal{P}$ & $\mathcal{H}$ & $\mathcal{I}$ 
                    & $\mathcal{P}$ & $\mathcal{H}$ & $\mathcal{I}$ \\
                    \midrule
                    \multirow{2}{*}{\textbf{OOD-1}}
                    & {\tt newstest-19} & 42.9 & 53.1 & 58.5 & 45.5 & 54.6 & 55.7 & 51.0 & 53.4 & \textbf{70.5} \\
                    & {\tt newstest-20} & 35.9 & 50.8 & 63.4 & 40.6 & 54.0 & 61.2 & 51.6 & 53.2 & \textbf{78.1} \\
                    \midrule
                    \multirow{2}{*}{\textbf{OOD-2}}
                    & {\tt khresmoi-dev} & 38.1 & 51.8 & 67.2 & 43.6 & 57.2 & 63.4 & 50.4 & 51.1 & \textbf{77.9} \\
                    & {\tt khresmoi-test} & 39.4 & 53.8 & 67.6 & 44.4 & 58.5 & 63.4 & 55.5 & 54.9 & \textbf{81.2} \\
                    \midrule
                    \multirow{2}{*}{\textbf{OOD-3}}
                    & {\tt mtnt-2019} & 66.0 & \textbf{72.2} & 64.4 & 67.0 & 72.0 & 61.9 & 70.4 & 72.0 & 71.4 \\
                    & {\tt kftt} & 31.9 & 33.8 & 47.0 & 32.6 & 35.8 & 40.8 & 27.3 & 34.8 & \textbf{54.7} \\
    			\bottomrule
    		\end{tabular}
			\label{tab:detection-performance}
		\end{center}
	\end{minipage}
\end{table}

\subsubsection{Estimating BERTScores in Machine Translation}
\label{sssec:bertscores-mt}

%
Table \ref{tab:BERTScore-performance} directly compares the rank correlation between model confidence/proxy scores and sentence BERTScore performance. We include proxies with attentive pooling as this is a more challenging task. These suggest that training NAPs directly on performance metrics provides a better predictor of a system's performance than using information-theoretic metrics such as confidence and entropy.

\begin{table}[h!]
	\centering{}
	\begin{minipage}[t]{1.0\textwidth}%
		\begin{center}
                \caption{Spearman Rank correlation score between model confidence/entropy and the model BERTScore (fine-tuned T5 Large). The NAPs were trained to predict this score directly. Averaged over 3 runs. Standard deviations in the order of $\pm 2.0$.}
			\vspace{1mm}
			\footnotesize
    		\begin{tabular}{ll|cc|ccc|ccc}
    			\toprule
                    \multirow{2}{*}{\textbf{Split}} & \multirow{2}{*}{\textbf{Dataset}} & \multicolumn{2}{c|}{\textbf{T5 Large}} & \multicolumn{3}{c|}{\textbf{NAP}} & \multicolumn{3}{c}{\textbf{NAP w/ Attention}} \\
                    & & $\mathcal{P}$ & $\mathcal{H}$
                    & {\tt S} & {\tt B} & {\tt L}
                    & {\tt S} & {\tt B} & {\tt L} \\
                    \midrule
                    \multirow{2}{*}{\textbf{ID}}
                    & {\tt iwslt-2017} & 16.6 & 41.6 & 42.0 & 43.7 & 44.9 & 42.5 & 44.4 & \textbf{45.6} \\
                    & {\tt ted-iwslt-2016} & 11.6 & 37.3 & 35.8 & 36.3 & 37.3 & 35.7 & 37.0 & \textbf{38.1} \\
                    \midrule
                    \multirow{2}{*}{\textbf{OOD-1}}
                    & {\tt newstest-19} & 32.9 & \textbf{39.3} & 34.3 & 36.7 & 37.6 & 34.7 & 37.1 & 39.2 \\
                    & {\tt newstest-20} & 34.2 & 38.3 & 38.6 & 38.7 & \textbf{39.6} & 38.9 & 39.0 & 39.3 \\
                    \midrule
                    \multirow{2}{*}{\textbf{OOD-2}}
                    & {\tt khresmoi-dev} & 41.4 & \textbf{45.5} & 40.8 & 43.1 & 44.7 & 41.3 & 42.3 & 44.8 \\
                    & {\tt khresmoi-test} & 42.9 & 46.1 & 42.0 & 46.5 & 45.5 & 42.3 & \textbf{47.8} & 45.2 \\
                    \midrule
                    & {\tt average} & 29.9 & 41.3 & 38.9 & 40.8 & 41.6 & 39.2 & 41.3 & \textbf{42.0} \\
    			\bottomrule
    		\end{tabular}
			\label{tab:BERTScore-performance}
		\end{center}
	\end{minipage}
\end{table}

\begin{figure}[h!]
    \centering
    \includegraphics[width=0.90\textwidth]{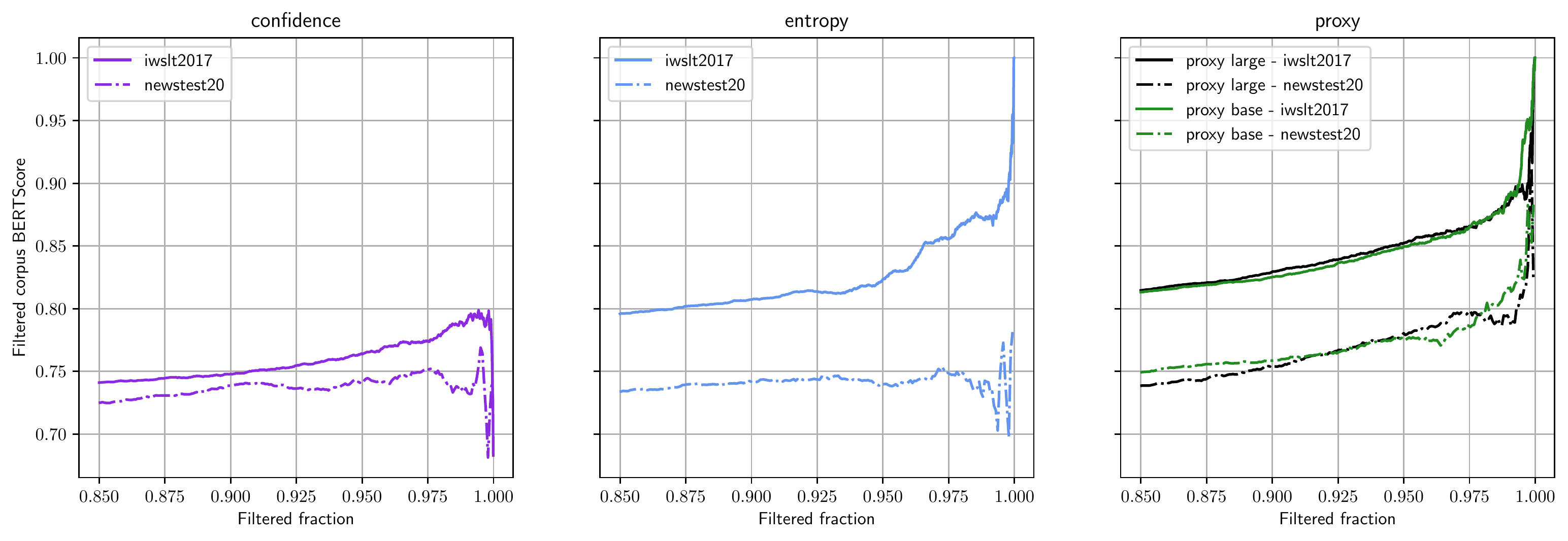}
    \caption{Measuring the BERTScore performance of T5 Large (primary) model on a filtered dataset when removing the worst examples according to some metric: (1) confidence, (2) entropy or (3) proxy outputs. The proxy was trained on the BERTScores of the primary model (on the training set).}
    \label{fig:BERTScore-filtering}
\end{figure}

Dataset filtering is an alternative approach to evaluating the quality of uncertainty or proxy estimates, with emphasis on the highest-performing examples.
A well-suited predictor of performance will show a monotonic increase in filtered dataset performance, as examples for which a low performance is predicted are removed.
Figure \ref{fig:BERTScore-filtering} shows this desired behaviour is best achieved with NAPs (equipped with attention pooling) that are directly trained to predict BERTScores of the primary model, in both an ID and OOD dataset. Entropy produced by the model itself is promising on the ID dataset but fails on OOD since the performance does not increase as we filter more inputs. 
Failure to reproduce these trends from uncertainty estimates of the primary model output suggests over-confidence \cite{calibration} in low-performing examples.

\begin{figure}[h!]
     \centering
     \hfill
     \begin{subfigure}[b]{0.400\textwidth}
         \centering
         \includegraphics[width=\textwidth]{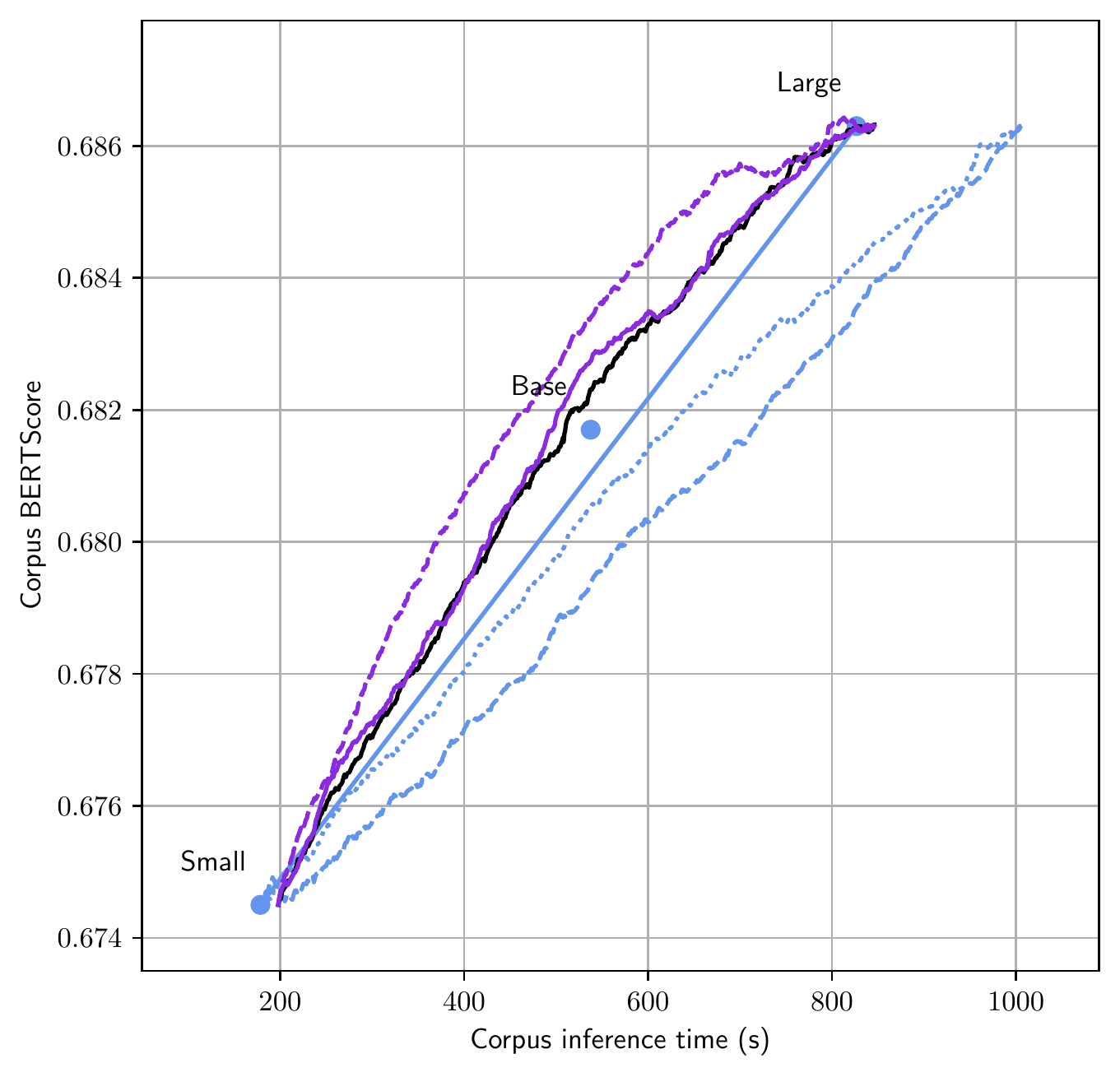}
         \caption{IWSLT 2017}
         \label{fig:resource-optimization-BERTScores-v1}
     \end{subfigure}
     \hfill
     \begin{subfigure}[b]{0.400\textwidth}
         \centering
         \includegraphics[width=\textwidth]{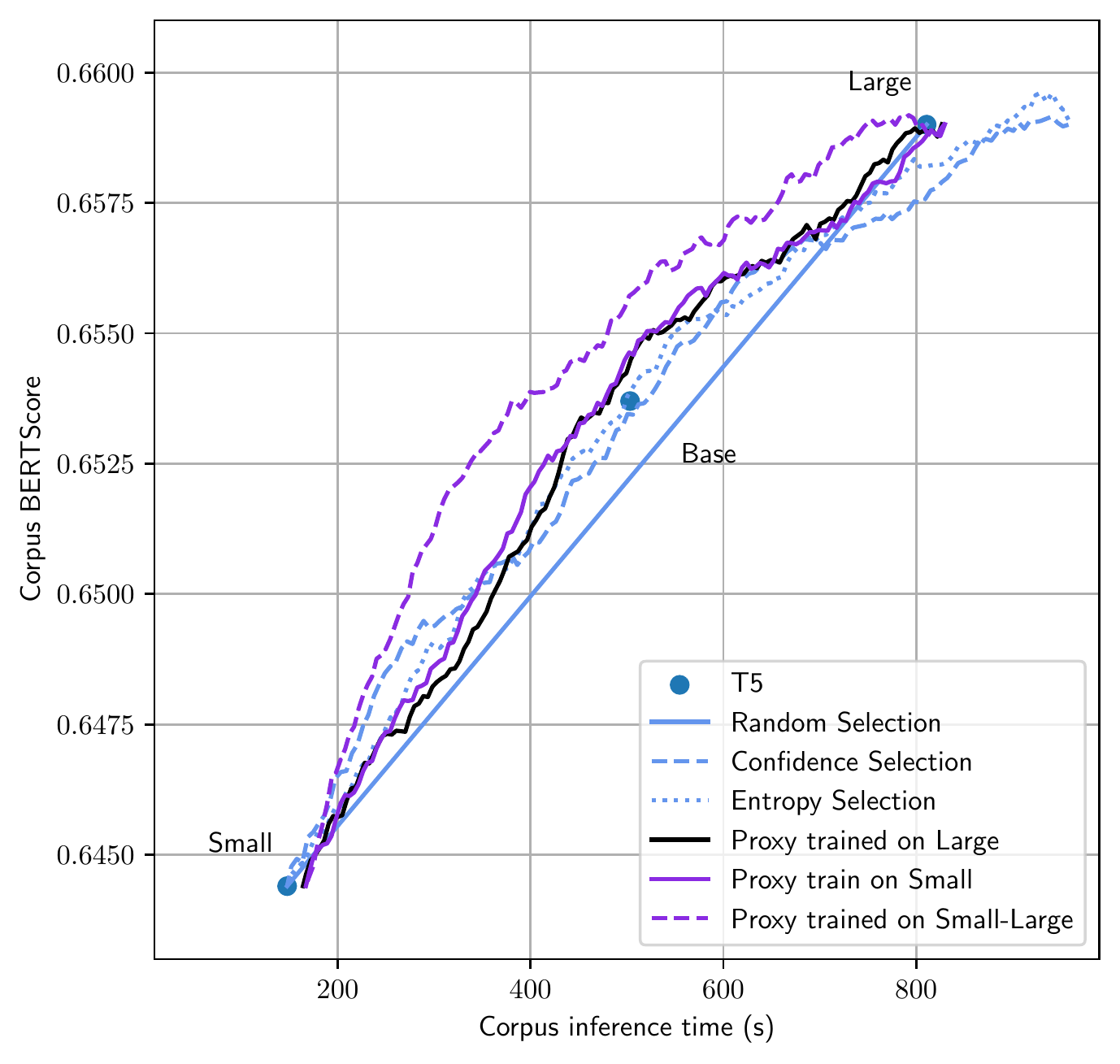}
         \caption{Newstest 20}
         \label{fig:resource-optimization-BERTScores-v2}
     \end{subfigure}
     \hfill
    \caption{Resource allocation: Measuring the overall BERTScore performance and inference time when distributing inputs between a T5 Small and Large according to some metric.}
    \label{fig:resource-optimization-BERTScores}
\end{figure}

Figure \ref{fig:resource-optimization-BERTScores} shows results for resource allocation, where examples are allocated to either a T5 Small or Large based on whether a performance-based related metric is above or below a threshold. Depending on the fraction allocated to the larger system, different levels of overall inference time and performance are achieved.
As expected from the dataset filtering results, proxy outputs can better predict instances for which the small model will perform poorly and it does so with a minuscule time cost. 
By contrast, relying on the output of the small model itself to decide whether the large model is required causes serious delays due to the time spent decoding, delays that the NAP preempts.
The best performance was achieved by NAPs trained on the \textit{difference} in BERTScore between the two available systems. The aim of this difference metric is to assign to the large model, examples for which we expect a maximal \textit{increase} in performance. Obtaining such a difference metric using the original models would defeat the whole purpose of resource optimization.
Finally, it is possible to be more efficient or better performing than a T5 Base using this deferral system while matching its performance or efficiency respectively.

%
%
%
%
%
%
%

\subsection{Estimating WERs in Automatic Speech Recognition}

We repeat experiments from Section \ref{sssec:bertscores-mt} using pretrained Whisper models from Hugging Face \cite{huggingface} on the LibriSpeech corpus \cite{librispeech}. We will by default use greedy decoding as opposed to beam-search since it was found to be robust enough \cite{whisper}. Table \ref{tab:whisper-speed} shows real-time factors (RTFs) demonstrating the inference efficiency of NAPs which do not require a decoder. Compared to greedy ($B = 1$) decoding of Whisper Large-V2, medium and large-sized NAPs are 43 and 33 times faster, respectively.

\begin{table}[h!]
	\centering{}
	\begin{minipage}[t]{1.0\textwidth}%
		\begin{center}
                \caption{Real-time Factors (RTFs). The proxy performs an encoder forward pass \& (pretrained) Whisper models perform beam search. Evaluated on {\tt test.other} using an NVIDIA A100. Corpus WER performance computed for the {\tt B = 1} setting which is used in downstream tasks.}
			\small
    		\begin{tabular}{r|ccc|c}
    			\toprule
                    \multirow{2}{*}{{\tt Model}} & \multicolumn{3}{c|}{{\tt Whisper Models}} & \multirow{2}{*}{\tt NAP} \\
                    & 
                    {\tt B = 1} & 
                    {\tt B = 5} & 
                    {\tt \%WER} 
                    \\
                    \midrule
                    {\tt Small} & 0.0480 & 0.0507 & 7.62 & 0.0014 \\
                    {\tt Medium} & 0.0722 & 0.1075 & 6.26 & 0.0024 \\
                    {\tt Large-V2} & 0.1029 & 0.1625 & 5.16 & 0.0031 \\
    			\bottomrule
    		\end{tabular}
			\label{tab:whisper-speed}
		\end{center}
	\end{minipage}
\end{table} 

%
%
%

Table \ref{tab:wer-performance} recreates the prior success of proxies in imitating model performance, in this case, sentence-level WER. Furthermore, since Whisper encoders pad all inputs to 30s, including an attention pooling layer can discount the padding and significantly improve performance. The following experiments will use the medium-sized NAP with attention pooling as default since it was found to have similar performance to its larger counterpart on the development sets but with a 23\% smaller RTF.

\begin{table}[h!]
	\centering
	\begin{minipage}[t]{1.0\textwidth}%
		\begin{center}
                \caption{Pearson correlation between Whisper Large-V2 confidence/entropy and sentence WER. The NAPs were trained to predict WER directly. Standard deviations in the order of $\pm 1.0$.}
			\vspace{1mm}
			\footnotesize
    		\begin{tabular}{l|cc|ccc|ccc}
    			\toprule
                    \multirow{2}{*}{\textbf{Dataset}} & \multicolumn{2}{c|}{\textbf{Whisper Large-V2}} & \multicolumn{3}{c|}{\textbf{NAP}} & \multicolumn{3}{c}{\textbf{NAP w/ Attention}} \\
                    & \hspace{3mm} $\mathcal{P}$ \hspace{3mm} & \hspace{3mm} $\mathcal{H}$ \hspace{3mm} 
                    & {\tt S} & {\tt M} & {\tt L}
                    & {\tt S} & {\tt M} & {\tt L} \\
                    \midrule
                    {\tt test.clean} & 13.3 & 16.8 & 32.4 & 36.3 & 33.9 & 43.9 & \textbf{49.7} & 47.2 \\
                    {\tt test.other} & 51.9 & 60.1 & 38.0 & 42.4 & 43.8 & 49.8 & 59.0 & \textbf{61.5} \\
    			\bottomrule
    		\end{tabular}
			\label{tab:wer-performance}
		\end{center}
	\end{minipage}
\end{table}

\begin{figure}[h!]
    \centering
    \includegraphics[width=0.90\textwidth]{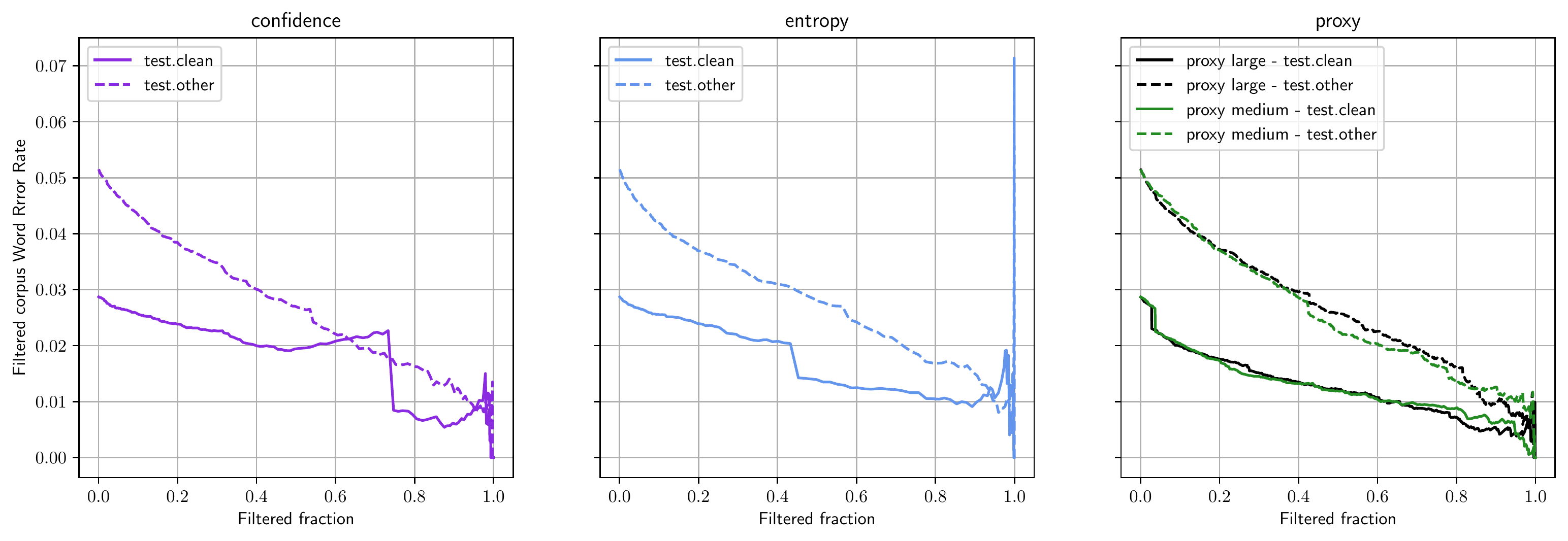}
    \vspace{-1.5mm}
    \caption{Measuring the corpus WER performance of Whisper Large-V2 (primary) on a filtered dataset when removing the worst examples according to some metric: (1) confidence, (2) entropy or (3) proxy outputs. The proxy was trained on the WERs of the primary model (on the training set).}
    \label{fig:wer-filtering}
\end{figure}

Figure \ref{fig:wer-filtering} shows the filtered corpus WER of {\tt test.clean} and {\tt test.other} when removing the worst examples according to model confidence/entropy or proxy outputs. While all are successful on {\tt test.other}, sequence-level confidence and entropy significantly suffer on {\tt test.clean} showing increasing corpus WER in certain regions when supposedly removing bad examples, a sign of over-confidence. This failure on {\tt test.clean} could have been somewhat predicted by the small correlations in Table \ref{tab:wer-performance} while NAPs with attention show a significantly better correlation performance with sentence WER.

\begin{figure}[h!]
    \centering
    \includegraphics[width=0.90\textwidth]{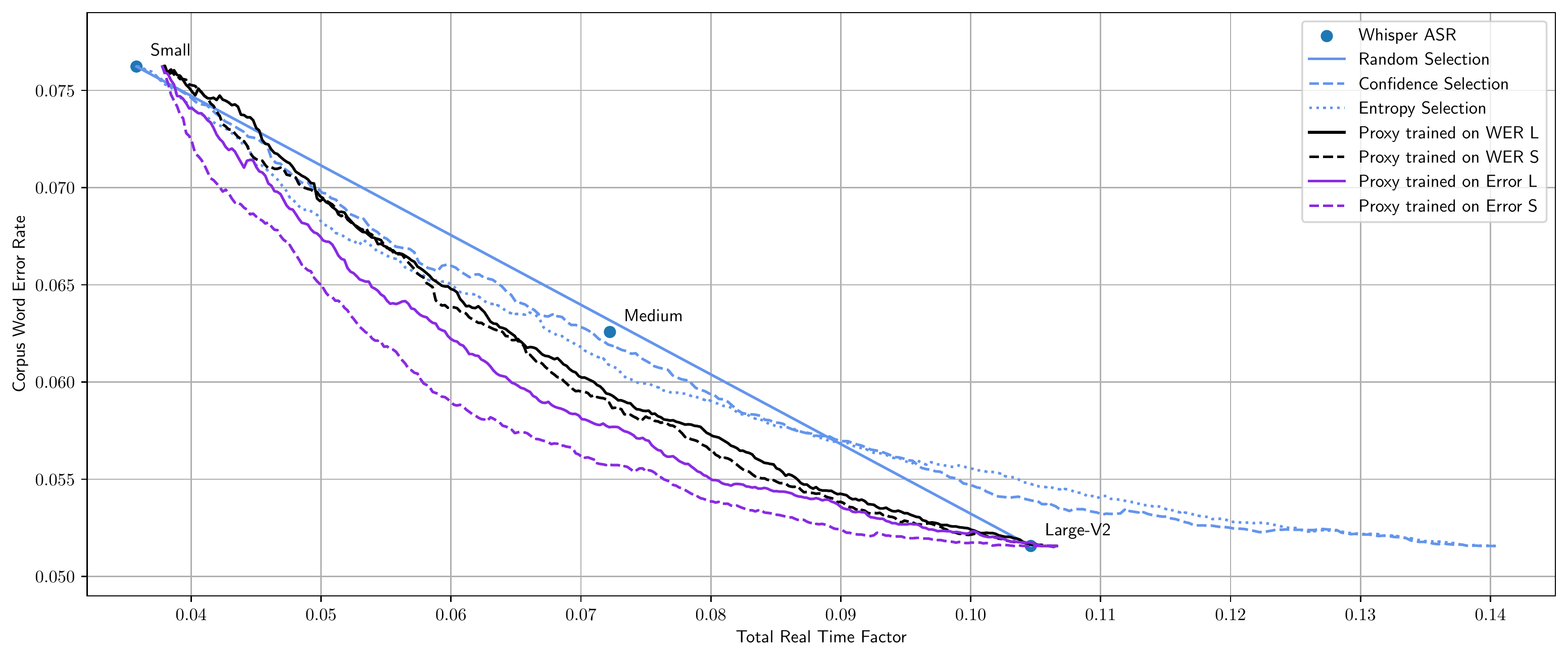}
    \vspace{-1.5mm}
    \caption{Resource allocation: Measuring the overall corpus WER performance and RTF when allocating inputs between a Whisper Small and Large-V2 according to some metric.}
    \label{fig:resource-optimization-wer}
\end{figure}

Figure \ref{fig:resource-optimization-wer} shows results for resource allocation, where examples are allocated to a Whisper Small or Large-V2 based on some performance-based related metric. 
Again, deferral systems using NAPs (with attention) significantly outperform decoder uncertainty-based selection schemes. 
In fact, the best-performing NAP here was one trained on the number of errors in a transcription, rather than the WER. This is simply because the ordinate in Figure \ref{fig:resource-optimization-wer} is the corpus WER, rather than the average sentence WER. This is proportional to the error count in the whole corpus, making this a more suitable optimization target.
Finally, we note that resource optimisation by training a proxy to predict a difference in WER or errors is not presented here. Since the Whisper Small and Large-V2 make the same number of word errors in approximately 75\% of examples on the training set, training a proxy on such a sparse label set is difficult.

\begin{wraptable}{r}{62mm}
    \centering
    \vspace{-4.2mm}
    \caption{Columns show (1) corpus WER performance of various deferral systems operating at the same RTF as Whipser Medium and (2) the RTF when operating at the same WER as Whipser Medium.}
    \footnotesize
    \begin{tabular}{l|c|c}
        \toprule
        Selection & WER & RTF \\
        \midrule
        Whisper Medium & 6.26 & 0.0722 \\ 
        \midrule
        Confidence Selection & 6.19 & 0.0707 \\ 
        Entropy Selection & 6.09 & 0.0677 \\ 
        \midrule
        Proxy trained on WER L & 5.94 & 0.0645 \\ 
        Proxy trained on WER S & 5.89 & 0.0640 \\ 
        Proxy trained on Error L & 5.77 & 0.0596 \\ 
        Proxy trained on Error S & 5.57 & 0.0534 \\ 
        \bottomrule
    \end{tabular}
    \label{tab:matched-settings}
    \vspace{-11mm}
\end{wraptable}
Finally, Table \ref{tab:matched-settings} shows the WER or RTF of various deferral systems (between Whisper Small and Large-V2) when operating at the Whisper Medium RTF or WER respectively. The best deferral system, a NAP trained on the number of errors of Whisper Small, reduces WER by 11\% while matching the inference speed of Whisper Medium. For the same WER performance, this system can reduce the RTF by 26\%.

\section{Conclusion}

For many downstream sequence-to-sequence tasks, only attributes of the output sequence are needed, and not the output itself. 
In this paper, we propose a simple, highly efficient, framework for directly estimating scalar sequence-level attributes, Non-Autoregressive Proxies (NAPs). These lightweight models  completely bypass the expensive autoregressive decoding process, only making use of an encoder stage. 
We show that NAPs can learn information-theoretic uncertainties as well as performance
metrics, such as BERTScores for MT or WERs for ASR, in terms of both mimicing attribute score  ranks and the impact on downstream tasks. For MT systems they outperform a deep ensemble on OOD detection with an order of magnitude higher inference speed. Furthermore, NAPs are able to outperform predictive uncertainty on downstream tasks such as data filtering and resource optimization on both ASR and MT tasks.

\clearpage
\AtNextBibliography{\small}
\printbibliography

\clearpage

\title{Who Needs Decoders? Efficient Estimation of Sequence-level Attributes\\(Supplementary Material)}
\author{%
  Yassir Fathullah, Puria Radmard, Adian Liusie, Mark J. F. Gales\\
  Engineering Department, University of Cambridge\\
  \texttt{\{yf286, pr450, al826, mjfg100\}@cam.ac.uk} \\
}

\maketitle

\appendix
\section{Experimental Configuration}

This section will describe the experimental setup of all experiments. Details about datasets, models, and training hyperparameters and evaluation are provided. Hugging Face was used extensively for all experiments in terms of loading various pretrained models, corresponding tokenizers and processed datasets.

\subsection{Machine Translation}
\label{ssec:mt}

\subsubsection{Datasets}
\label{sssec:mt-data}

In Table \ref{tab:datasets} we report information about the datasets we use for training and evaluation. Note that we use the T5 \cite{t5} approach for English-to-German tokenization meaning that we prepend the following prompt to all inputs "translate English to German:" prior to tokenization. We use the {\tt iwslt-2017} training set for finetuning T5 systems on spoken language translation and evaluate on the corresponding test set. We furthermore use the in-domain (ID) spoken language test set and out-of-domain news commentary (OOD-1), medical data (OOD-2) and a final mixed category of noisy text and Japanese articles (OOD-3) for downstream tasks.

\begin{table}[h!]
	\centering{}
	\begin{minipage}[t]{1.0\textwidth}%
		\begin{center}
			\caption{Dataset statistics post tokenization.}
			\def\arraystretch{1.20}
			\begin{adjustbox}{center}
    			\begin{tabular}{cc|c|cc}
    				\toprule
    				\multirow{2}{*}{\textbf{Split}} & 
    				\multirow{2}{*}{\textbf{Dataset}} & 
    				\multirow{2}{*}{\textbf{\#Sequences}} & 
    				\multicolumn{2}{c}{\textbf{\#Tokens/Sequence}} \\
    				& & &
    				\hspace{4mm} {\tt src} \hspace{4mm} & 
    				\hspace{4mm} {\tt ref} \hspace{4mm} \\
    				\midrule
    				\multirow{1}{*}{\textbf{Training}} & 
                        \multirow{3}{*}{{\tt iwslt-2017}} 
                        & 206,112 & 29.1 & 28.5 \\
                        \multirow{1}{*}{\textbf{Validation}} & 
                        & 888 & 31.9 & 32.7 \\
                        \multirow{1}{*}{\textbf{Evaluation}} & 
                        & 8,079 & 27.8 & 27.5 \\
    				\midrule
                        \multirow{1}{*}{\textbf{ID}}
    				& {\tt ted-iwslt-2016} & 3,662 & 46.4 & 54.2 \\
                        \midrule
    				\multirow{2}{*}{\textbf{OOD-1}}
    				& {\tt newstest-19}     & 1,997 & 35.3 & 39.7 \\
    				& {\tt newstest-20}     & 1,418 & 49.1 & 61.6 \\
    				\midrule
    				\multirow{2}{*}{\textbf{OOD-2}}
    				& {\tt khresmoi-dev}    & 500 & 33.7 & 38.6 \\
    			    & {\tt khresmoi-test}   & 1,000 & 34.7 & 40.4 \\
                        \midrule
    				\multirow{2}{*}{\textbf{OOD-3}}
    			    & {\tt mtnt-2019}       & 1,392 & & - \\
    			    & {\tt kftt}            & 1,160 & & - \\
    				\bottomrule
    			\end{tabular}
    		\end{adjustbox}
			\label{tab:datasets}
		\end{center}
	\end{minipage}
\end{table} 
\clearpage

\subsubsection{Models}
\label{sssec:mt-models}

All experiments use the T5 model. In Table \ref{tab:parameters} we report parameter counts of various models. The T5 is an encoder-decoder model with a language model head which predicts a probability mass function over every token in the output sequence. The proxy model consists of a T5 encoder and a head for predicting uncertainty. The parameter counts below are reported for a proxy with an average pooling layer; an attentive pooling layer would add some parameters. Note, although the embedding layer is expensive parameter-wise, it is extremely fast inference-wise since it is equivalent to a lookup table.

\begin{table}[h!]
	\centering{}
	\begin{minipage}[t]{1.0\textwidth}%
		\begin{center}
                \caption{Parameter counts of models. NAPs do not use a decoder during inference.}
			\def\arraystretch{1.20}
			\begin{adjustbox}{center}
    			\begin{tabular}{r|cccc|c}
    				\toprule
                        {\tt Model} & {\tt Embeddings} & {\tt Encoder} & {\tt Decoder} & {\tt Head} & {\tt Total} \\
                        \midrule
                        {\tt T5 Small} & \multirow{2}{*}{16.4M} & \multirow{2}{*}{35.3M} & 41.6M & 16.4M & 60.5M \\
                        {\tt NAP Small} & & & - & 5.2M & 40.6M \\
                        \midrule
                        {\tt T5 Base} & \multirow{2}{*}{24.7M} & \multirow{2}{*}{109.6M} & 137.9M & 24.7M & 222.9M \\
                        {\tt NAP Base} & & & - & 11.8M & 121.4M \\
                        \midrule
                        {\tt T5 Large} & \multirow{2}{*}{32.9M} & \multirow{2}{*}{334.9M} & 435.6M & 32.9M & 737.7M \\
                        {\tt NAP Large} & & & - & 20.9M & 355.9M \\
    				\bottomrule
    			\end{tabular}
    		\end{adjustbox}
			\label{tab:parameters}
		\end{center}
	\end{minipage}
\end{table}

\subsubsection{Finetuning T5 Models}
\label{sssec:mt-finetuning}

All T5 models were finetuned on the IWSLT-2017 \cite{iwslt2017} training set and evaluated on several ID and OOD datasets using both SacreBLEU \cite{sacrebleu} and BERTScore (BS) \cite{bertscore}, see Table \ref{tab:baseline-bleu}. We set the beam size to 12 and used a length penalty of 0.60.

The learning rate was fixed to 0.0001 and the batch size was selected to maximise GPU memory usage on a single NVIDIA A100 SXM4 80GBs. The performance was tracked on the validation set 10 times per epoch and training was terminated when performance did not improve for a whole epoch.

\begin{table}[h!]
	\centering{}
	\begin{minipage}[t]{1.0\textwidth}%
		\begin{center}
                \caption{SacreBLEU and BERTScore performance of finetuned T5 models.}
                \def\arraystretch{1.20}
    		\begin{tabular}{cc|cc|cc|cc}
    			\toprule
                    \multirow{2}{*}{\textbf{Split}} & \multirow{2}{*}{\textbf{Dataset}} & \multicolumn{2}{c|}{\textbf{Small}} & \multicolumn{2}{c|}{\textbf{Base}} & \multicolumn{2}{c}{\textbf{Large}} \\
                    & & {\tt BLEU} & {\tt BS} & {\tt BLEU} & {\tt BS} & {\tt BLEU} & {\tt BS} \\
                    \midrule
                    \multirow{2}{*}{\textbf{ID}}
                    & {\tt iwslt-2017} & 
                    32.0 & 67.4 & 
                    33.8 & 68.2 & 
                    34.3 & 68.6 \\
                    & {\tt ted-iwslt-2016} & 
                    30.9 & 65.2 & 
                    31.9 & 65.9 & 
                    32.3 & 66.3 \\
                    \midrule
                    \multirow{2}{*}{\textbf{OOD-1}}
                    & {\tt newstest-19} & 
                    37.3 & 68.0 & 
                    38.9 & 69.8 & 
                    38.9 & 69.9 \\
                    & {\tt newstest-20} & 
                    29.4 & 64.4 & 
                    30.8 & 65.4 & 
                    31.4 & 65.9 \\
                    \midrule
                    \multirow{2}{*}{\textbf{OOD-2}}
                    & {\tt khresmoi-dev} &
                    27.1 & 68.9 & 
                    29.2 & 70.7 & 
                    29.4 & 70.7 \\
                    & {\tt khresmoi-test} & 
                    27.4 & 68.0 & 
                    30.0 & 70.2 & 
                    30.2 & 70.3 \\
    			\bottomrule
    		\end{tabular}
			\label{tab:baseline-bleu}
		\end{center}
	\end{minipage}
\end{table}

The table shows that increasing the size of the T5 model improves performance on the ID datasets. Surprisingly the performance gap between the base and large configuration is very small for most OOD datasets, showing that the base model is particularly effective despite being more than a third of the size.

\subsubsection{Training Non-Autoregressive Proxies}
\label{sssec:mt-proxy}

We generated scores (uncertainty or BERTScore) from finetuned T5 Large models and used them to train NAP models. We used the smooth and differentiable extension to the Spearman Rank loss function \cite{sorting} which requires a hyperparameter controlling the level of smoothing. This hyperparameter was set to 0.000001 in all experiments. Similar to the section above, all experiments used a learning rate of 0.0001, maximised batch size and training was stopped when performance did not improve after an epoch.

\subsubsection{Estimating Uncertainties in Machine Translation}
\label{sssec:mt-unc}

The experiments in this section used the training set of IWSLT-2017 and followed Setup 1, see Figure 1a. The main T5 model produced sequence-level confidence or entropy uncertainty estimates under the reference sequence. The NAP model was then trained to capture this uncertainty. We could have also opted to generate sequence-level uncertainties using Setup 2 (see Figure 1b) but the quality of the uncertainties then depends on the quality of the decoded hypotheses. If we work with unlabelled datasets, we can always revert back to Setup 2 and train our proxy to imitate the uncertainties of the free-running hypotheses.

The performance of the uncertainty estimation NAP was then compared to the main model in two ways. We first computed the Spearman Rank correlation between the NAP output and the main model which was given the reference output. The second and more important evaluation was based on out-of-distribution detection. For this task, we took one in-domain dataset (IWSLT-2017 test set) and compared it with one of the out-of-distribution datasets mentioned above. We sought low uncertainties for the ID dataset and high uncertainties for the OOD dataset. We used the AUROC \cite{auroc} metric for measuring detection performance, where 50\% represents a fully random system.

\subsubsection{Estimating BERTScores in Machine Translation}
\label{sssec:mt-bert}

We decoded a finetuned T5 Large system (with a beam of $B = 12$ and length-penalty of 0.60) on the IWSLT-2017 training set. The decoded outputs were used to compute the BERTScore for each instance, following Setup 2. The NAP was then trained using the exact same hyperparameters as the above section.

Similar to the section above, the outputs of the NAP were first compared with the main model on several unseen datasets. Following, we evaluated the performance of this system on two downstream tasks. First, we took a dataset and filtered out samples with the lowest estimated BERTScore and computed the average BERTScore of the remaining samples. For a well-performing metric, we expect the average BERTScore of the remaining samples to increase monotonically. 

Next, we also performed a resource optimization task in which we used the NAP output to decide whether an input should be passed to a smaller (T5 Small) or larger more robust (T5 Large) system. When a proxy output is above a threshold, the input was passed to a smaller system and otherwise to the slower and larger system. The threshold therefore had a large impact on the performance and inference speed of the two model system. By selecting different thresholds, different operating points were achieved. A good system would achieve better performance while deferring as few samples as possible to the slower system. 

Furthermore, we also train a NAP to predict the BERTScore difference between the two models in the deferral system. This can be motivated by a simple example: Consider two different models, a smaller $\mathcal{M}_1$ and a larger more robust $\mathcal{M}_2$. Given two different inputs $\bm x_1$ and $\bm x_2$ the two models achieve the following BERTScores:

\begin{table}[h!]
	\centering{}
	\begin{minipage}[t]{1.0\textwidth}%
		\begin{center}
                \caption{Simple example.}
			\def\arraystretch{1.20}
			\begin{adjustbox}{center}
    			\begin{tabular}{c|ccc}
                        \toprule
                        & $\mathcal{M}_1$ & $\mathcal{M}_2$ & $\mathcal{M}_2 - \mathcal{M}_1$\\
                        \midrule
                        $\bm x_1$ & 0.70 & 0.90 & 0.20 \\
                        $\bm x_2$ & 0.50 & 0.40 & -0.10 \\
                        \bottomrule
    			\end{tabular}
    		\end{adjustbox}
			\label{tab:simple-example}
		\end{center}
	\end{minipage}
\end{table} 

Clearly, the first input is easier to handle since both models achieve higher BERTScores with $\mathcal{M}_2$ being stronger. If we performed an allocation based on the isolated performance of a single model itself, we would give the simpler example $\bm x_1$ to the smaller model $\mathcal{M}_1$ and the harder input $\bm x_2$ to the larger model achieving an average performance of 0.55 BERTScore. \textbf{However}, if we instead perform an allocation based on the performance difference, and refer samples to the stronger model $\mathcal{M}_2$ where it dominates (and vice versa), we would allocate $\bm x_1$ to model $\mathcal{M}_2$ and $\bm x_2$ to model $\mathcal{M}_1$ achieving an average score of 0.70. This shows that an allocation system should focus on the performance difference of the relevant metric.

\subsection{Automatic Speech Recognition}
\subsubsection{Datasets}
\label{sssec:asr-data}

Table \ref{tab:whisper-datasets-v1} includes information about the LibriSpeech corpus \cite{librispeech}. The number of words per sequence is computed based on the Whisper text normalization scheme. In this task, we do not finetune the ASR models and do not use any out-of-domain datasets. Instead, we focus on the noisy {\tt validation.other} and {\tt test.other} sets.

\begin{table}[h!]
	\centering{}
	\begin{minipage}[t]{1.0\textwidth}%
		\begin{center}
			\caption{Dataset statistics.}
			\def\arraystretch{1.20}
			\begin{adjustbox}{center}
    			\begin{tabular}{c|c|c}
    				\toprule
    				\multirow{2}{*}{\textbf{Dataset}} & 
    				\multirow{2}{*}{\textbf{\#Sequences}}      & 
    				\textbf{\#Words per} \\
    				& & \textbf{Sequence} \\
    				\midrule
                        \multirow{1}{*}{{\tt train.clean.100}} 
                        & 28,539 & 35.0 \\
                        \multirow{1}{*}{{\tt train.clean.360}} 
                        & 104,014 & 34.8 \\
                        \multirow{1}{*}{{\tt train.other.500}} 
                        & 148,688 & 32.7 \\
                        \midrule
                        \multirow{1}{*}{{\tt validation.clean}} 
                        & 2,703 & 20.3 \\
                        \multirow{1}{*}{{\tt validation.other}} 
                        & 2,864 & 18.0 \\
                        \midrule
                        \multirow{1}{*}{{\tt test.clean}} 
                        & 2,620 & 20.2 \\
                        \multirow{1}{*}{{\tt test.other}} 
                        & 2,939 & 18.0 \\
    				\bottomrule
    			\end{tabular}
    		\end{adjustbox}
			\label{tab:whisper-datasets-v1}
		\end{center}
	\end{minipage}
\end{table} 

\subsubsection{Models}
\label{sssec:asr-models}

In Table \ref{tab:whisper-parameters} we report parameter counts of various models. Whisper is an encoder-decoder model with a language model head which predicts a probability mass function over every token in the output sequence. The proxy model consists of a Whisper encoder and a head for predicting uncertainty. The parameter counts below are reported for a NAP with an average pooling layer; an attentive pooling layer would add some parameters.

\begin{table}[h!]
	\centering{}
	\begin{minipage}[t]{1.0\textwidth}%
		\begin{center}
                \caption{Parameter counts of models. NAPs do not use a decoder during inference.}
			\def\arraystretch{1.20}
			\begin{adjustbox}{center}
    			\begin{tabular}{r|ccc|c}
    				\toprule
                        {\tt Model} & {\tt Encoder} & {\tt Decoder} & {\tt Head} & {\tt Total} \\
                        \midrule
                        {\tt Whisper Small} & \multirow{2}{*}{88.1M} & 153.6M & 39.8M & 241.7M \\
                        {\tt NAP Small} & & - & 14.2M & 102.3M \\
                        \midrule
                        {\tt Whisper Medium} & \multirow{2}{*}{307.2M} & 456.6M & 53.1M & 763.9M \\
                        {\tt NAP Medium} & & - & 25.2M & 332.4M \\
                        \midrule
                        {\tt Whisper Large-v2} & \multirow{2}{*}{636.8M} & 906.5M & 66.4M & 1543.3M \\
                        {\tt NAP Large-v2} & & - & 39.3M & 676.1M \\
    				\bottomrule
    			\end{tabular}
    		\end{adjustbox}
			\label{tab:whisper-parameters}
		\end{center}
	\end{minipage}
\end{table}

\subsubsection{Training Non-Autoregressive Proxies}
\label{sssec:asr-proxy}

We generated sentence-level word error rates (WERs) from the Whisper Large-V2 model using greedy search. While it was found that a beam of $B = 5$ was the best-performing setting in the original work \cite{whisper}, this was only achieved using a highly non-standard decoding mechanism; simply using beam search with $B = 5$ actually degrades performance. Therefore, we opted for a simpler setup using greedy search, see Table \ref{tab:whisper-libri-greedy}.

\begin{table}[h!]
	\centering{}
	\begin{minipage}[t]{1.0\textwidth}%
		\begin{center}
			\caption{Baseline \%WER performance with greedy decoding.}
			\def\arraystretch{1.20}
			\begin{adjustbox}{center}
    			\begin{tabular}{l|ccc}
    				\toprule
    				\multirow{2}{*}{\textbf{Dataset}} & 
    				\multirow{2}{*}{\hspace{1.5mm} \textbf{Small} \hspace{1.5mm}} & 
    				\multirow{2}{*}{\textbf{Medium}} & 
    				\multirow{2}{*}{\textbf{Large-v2}} \\ \\
    				\midrule
                        {\tt validation.clean} & 3.70 & 2.69 & 2.48 \\
                        {\tt validation.other} & 7.35 & 5.46 & 4.96 \\
                        \midrule
                        {\tt test.clean} & 3.45 & 2.88 & 2.87 \\
                        {\tt test.other} & 7.62 & 6.26 & 5.16 \\
    				\bottomrule
    			\end{tabular}
    		\end{adjustbox}
			\label{tab:whisper-libri-greedy}
		\end{center}
	\end{minipage}
\end{table} 

When generating the sentence WERs on the training data of the LibriSpeech corpus, it was found that approximately half of all instances were correctly decoded. This would present problems for a ranking loss and we instead opted to train all NAP models using the Pearson correlation loss. Similar to the section above, all experiments used a learning rate of 0.0001, maximised batch size and training was stopped when performance did not improve after an epoch.

\subsection{Estimating WERs in Automatic Speech Recognition}
\label{sssec:asr-wer}

Following the exact same line of experiments as in Section \ref{sssec:mt-bert}. A NAP was trained to imitate the sentence-level WERs and was evaluated on two downstream tasks, filtering and resource allocation. Note that we train additional proxy systems to capture the total number of errors (instead of the error rate) since this is more aligned with the resource allocation task. The resource allocation was done between the Whisper Large-V2 and Whisper Small models. 

We are unable to train a system to capture the error difference for the resource allocation task since training the NAP was unstable. Approximately 74\% of all error differences on the training set were 0 making it a highly imbalanced dataset.

\section{Ablation Studies}

We run all of our ablation studies on capturing mutual information of a T5 Large ensemble on the machine translation task. The ensemble consists of three members.

\subsection{Choice of Loss Function}

All of the experiments in the main paper used a differentiable Spearman correlation coefficient ({\tt scc}) loss. This section explores alternative loss functions including mean absolute error ({\tt mae}), root mean squared error ({\tt rmse}) and pearson correlation coefficient ({\tt pcc}).

\begin{table}[h!]
	\centering{}
	\begin{minipage}[t]{1.0\textwidth}%
		\begin{center}
                \caption{Detection performance of NAPs using MI $\mathcal{I}$.}
                    \def\arraystretch{1.20}
    			\begin{tabular}{ll|cccc}
    				\toprule
                        \multirow{2}{*}{\textbf{Split}} & \multirow{2}{*}{\textbf{Dataset}} & \multicolumn{4}{c}{\textbf{NAP Large}} \\
                        & & {\tt mae} & {\tt rmse} & {\tt pcc} & {\tt scc} \\
                        \midrule
                        \multirow{2}{*}{\textbf{OOD-1}}
                        & {\tt newstest-19} & 67.3 & 66.9 & 69.6 & \textbf{70.5} \\
                        & {\tt newstest-20} & 74.9 & 73.6 & 76.0 & \textbf{78.1} \\
                        \midrule
                        \multirow{2}{*}{\textbf{OOD-2}}
                        & {\tt khresmoi-dev} & 77.9 & 78.2 & \textbf{79.1} & 77.9 \\
                        & {\tt khresmoi-test} & 80.5 & 81.0 & \textbf{81.5} & 81.2 \\
                        \midrule
                        \multirow{2}{*}{\textbf{OOD-3}}
                        & {\tt mtnt-2019} & 69.5 & 71.4 & \textbf{73.4} & 71.4 \\
                        & {\tt kftt} & 50.2 & 50.2 & 52.8 & \textbf{54.7} \\
                        \midrule
                        & {\tt average} & 70.1 & 70.2 & 72.1 & \textbf{72.3} \\
    				\bottomrule
    			\end{tabular}
			\label{tab:detection-ablation-loss-function}
		\end{center}
	\end{minipage}
\end{table} 

The correlation-based loss functions are consistently better than mean absolute and root mean squared error losses, possibly because the correlation losses do not require accurate prediction of the uncertainties, only their ordering.

\subsection{Predictor Architecture}

\begin{figure}[h!]
    \centering
    \hspace{5mm}
    \begin{subfigure}[b]{0.197\textwidth}
        \centering
        \includegraphics[width=\textwidth]{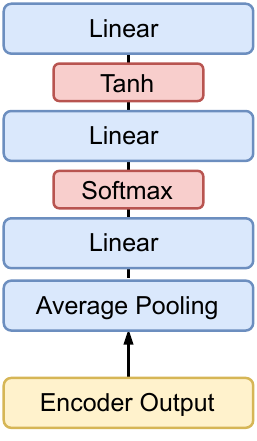}
        \caption{Standard.}
        \label{fig:proxy-head-avg}
    \end{subfigure}
    \hspace{15mm}
    \begin{subfigure}[b]{0.410\textwidth}
        \centering
        \includegraphics[width=\textwidth]{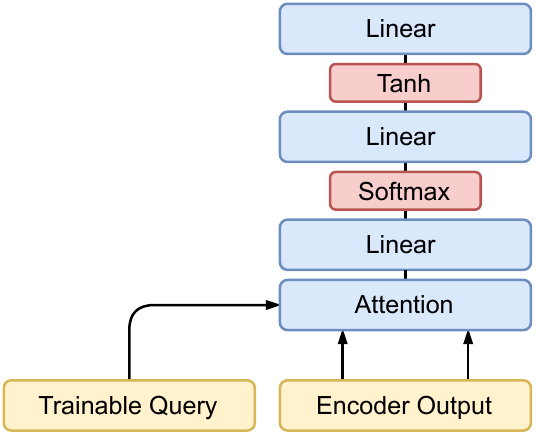}
        \caption{With attentive pooling.}
        \label{fig:proxy-head-attn}
    \end{subfigure}
    \hspace{5mm}
    \caption{The standard three-layer network is used on top of a non-autoregressive proxy. When average pooling the encoder output is restrictive, an attention layer is used instead with a trainable query.}
    \label{fig:proxy-heads}
\end{figure}

We also investigate the architecture, and specifically the activations of the MLP that is added on top of the NAP encoder, see Figure \ref{fig:proxy-heads}. In a toy example, we found that a two-layer (with tanh activation) network is better able to predict entropy scores from categorical predictions. This motivates using a three-layer network with an initial softmax activation to produce 'virtual' probabilities.

\begin{figure}[h!]
    \centering
    \begin{subfigure}[b]{1.00\textwidth}
        \centering
        \includegraphics[width=\textwidth]{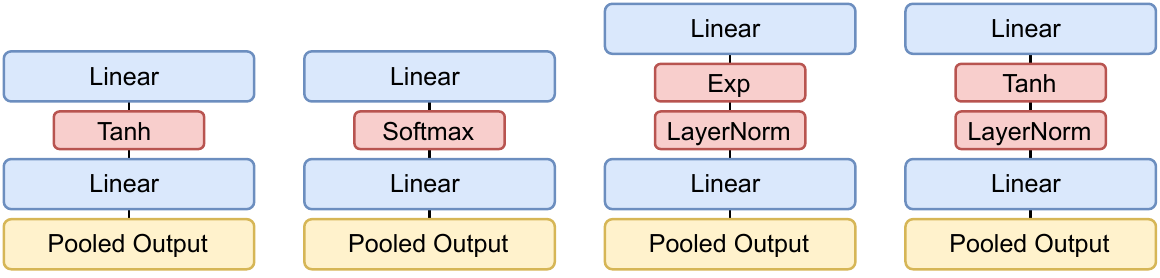}
        \caption{From left to right: \{{\tt 2L Tanh}, {\tt 2L SM}, {\tt 2L LN-Exp} \& {\tt 2L LN-Tanh}\}.}
        \vspace{4mm}
        \label{fig:proxy-head-2l}
    \end{subfigure}
    \vspace{4mm}
    \begin{subfigure}[b]{1.00\textwidth}
        \centering
        \includegraphics[width=\textwidth]{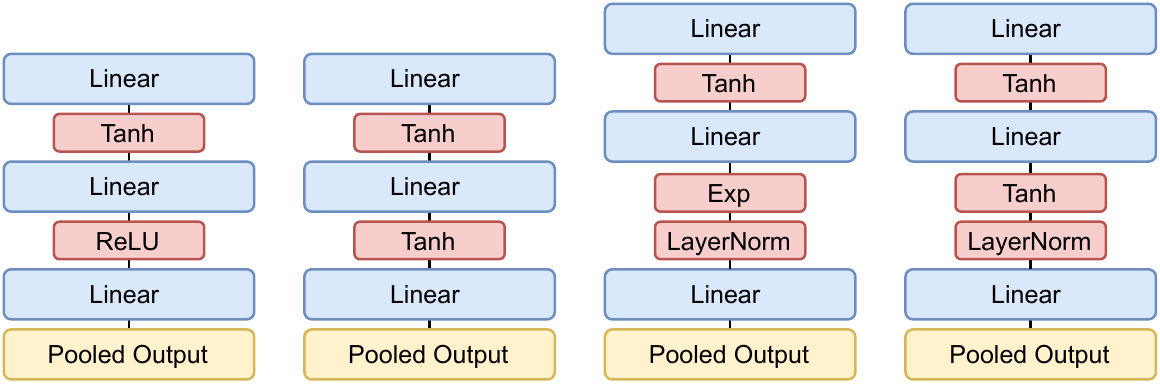}
        \caption{From left to right: \{{\tt 3L ReLU}, {\tt 3L Tanh}, {\tt 3L LN-Exp} \& {\tt 3L LN-Tanh}\}.}
        \label{fig:proxy-head-3l}
    \end{subfigure}
    \caption{Various configurations of proxy heads investigated.}
    \label{fig:proxy-heads-v2}
\end{figure}

This section also explores a range of different (parameter-matched) two-layer and three-layer MLPs with various activation functions, see Figure \ref{fig:proxy-heads-v2}.

Table \ref{tab:detection-ablation-predictor} shows the performance of various MLPs (with average pooling) in the out-of-distribution detection task. The two-layer and three-layer MLPs are parameter matched. The final model {\tt 3L SM} is the default MLP head used in all experiments. Clearly, the use of a softmax activation is extremely important for achieving the best possible performance. 

\begin{table}[h!]
	\centering{}
	\begin{minipage}[t]{1.0\textwidth}%
		\begin{center}
                \caption{Detection performance of NAPs using MI $\mathcal{I}$.}
                \def\arraystretch{1.20}
			\begin{adjustbox}{center}
    			\begin{tabular}{ll|cccccccc|c}
    				\toprule
                        \multirow{3}{*}{\textbf{Split}} & \multirow{3}{*}{\textbf{Dataset}} & \multicolumn{8}{c|}{\hspace{3mm} \textbf{NAP Large}} \\
                        & & {\tt 2L} & {\tt 2L} & {\tt 2L} & {\tt 2L} & {\tt 3L} & {\tt 3L} & {\tt 3L} & {\tt 3L} & {\tt 3L} \\
                        & & {\tt Tanh} & {\tt SM} & {\tt LN-Exp} & {\tt LN-Tanh} & {\tt ReLU} & {\tt Tanh} & {\tt LN-Exp} & {\tt LN-Tanh} & {\tt SM} \\
                        \midrule
                        \multirow{2}{*}{\textbf{OOD-1}}
                        & {\tt newstest-19} & 56.6 & 67.7 & 50.5 & 48.4 & 46.4 & 57.2 & 59.9 & 59.7 & \textbf{70.5} \\
                        & {\tt newstest-20} & 66.2 & 75.4 & 58.6 & 56.0 & 47.0 & 68.2 & 67.7 & 63.2 & \textbf{78.1} \\
                        \midrule
                        \multirow{2}{*}{\textbf{OOD-2}}
                        & {\tt khresmoi-dev} & 55.6 & 77.5 & 66.4 & 49.8 & 39.2 & 52.8 & 65.1 & 59.1 & \textbf{77.9} \\
                        & {\tt khresmoi-test} & 56.0 & 80.6 & 67.4 & 51.8 & 38.9 & 53.8 & 65.2 & 62.2 & \textbf{81.2} \\
                        \midrule
                        \multirow{2}{*}{\textbf{OOD-3}}
                        & {\tt mtnt-2019} & 54.1 & \textbf{71.6} & 48.4 & 52.6 & 63.4 & 47.8 & 61.4 & 50.6 & 71.4 \\
                        & {\tt kftt} & 55.2 & 50.4 & 55.9 & 52.0 & 43.0 & \textbf{62.0} & 58.1 & 44.8 & 54.7 \\
                        \midrule
                        & {\tt average} & 57.3 & 70.5 & 57.9 & 51.8 & 46.3 & 56.9 & 62.9 & 56.6 & \textbf{72.3} \\
    				\bottomrule
    			\end{tabular}
    		\end{adjustbox}
			\label{tab:detection-ablation-predictor}
		\end{center}
	\end{minipage}
\end{table} 

\subsection{Intermediate Outputs of Encoder}

It is not necessary to pick the final layer output as the input to the predictor MLP. One can use intermediate layer outputs as well. Previous work has found that using intermediate outputs can even improve upon a task \cite{hubert, bertscore}. Using intermediate layer outputs also leads to faster inference and lower parameter counts, see Table \ref{tab:intermediate-parameters}.

\begin{table}[h!]
	\centering{}
	\begin{minipage}[t]{1.0\textwidth}%
		\begin{center}
                \caption{Parameter counts and inference time of models on {\tt iwslt-2017}.}
			\def\arraystretch{1.20}
			\begin{adjustbox}{center}
    			\begin{tabular}{r|cccc|c}
    				\toprule
                        {\tt Layers} & {\tt Embeddings} & {\tt Encoder} & {\tt Head} & {\tt Total} & {\tt Inference Time} \\
                        \midrule
                        {\tt Default 24L} & 32.9M & 334.9M & 20.9M & 355.9M & 17.9s \\
                        {\tt 21L} & 32.9M & 289.2M & 20.9M & 310.1M & 15.3s \\
                        {\tt 18L} & 32.9M & 259.4M & 20.9M &  280.4M & 12.7s \\
                        {\tt 15L} & 32.9M & 221.7M & 20.9M & 242.7M & 9.9s \\
                        {\tt 12L} & 32.9M & 184.0M & 20.9M & 204.9M & 7.5s \\
    				\bottomrule
    			\end{tabular}
    		\end{adjustbox}
			\label{tab:intermediate-parameters}
		\end{center}
	\end{minipage}
\end{table} 

According to Table \ref{tab:detection-ablation-intermediate}, the performance of NAPs remains arguably consistent when utilizing intermediate outputs down until the 12th layer, where performance starts dropping. Therefore, it is possible based on this experiment to remove the top 9 layers of the T5 encoder reducing the total parameter count by 32\% and inference time by 45\% without notably sacrificing performance.

\begin{table}[h!]
	\centering{}
	\begin{minipage}[t]{1.0\textwidth}%
		\begin{center}
                \caption{Detection performance of NAPs using MI $\mathcal{I}$.}
                \def\arraystretch{1.20}
			\begin{adjustbox}{center}
    			\begin{tabular}{ll|ccccc}
    				\toprule
                        \multirow{2}{*}{\textbf{Split}} & \multirow{2}{*}{\textbf{Dataset}} & \multicolumn{5}{c}{\textbf{NAP Large}} \\
                        & & {\tt 24L} & {\tt 21L} & {\tt 18L} & {\tt 15L} & {\tt 12L} \\
                        \midrule
                        \multirow{2}{*}{\textbf{OOD-1}}
                        & {\tt newstest-19} & 70.5 & 68.7 & 69.1 & 68.6 & 68.1 \\
                        & {\tt newstest-20} & 78.1 & 77.0 & 77.1 & 76.0 & 75.4 \\
                        \midrule
                        \multirow{2}{*}{\textbf{OOD-2}}
                        & {\tt khresmoi-dev} & 77.9 & 78.5 & 77.2 & 77.0 & 76.4 \\
                        & {\tt khresmoi-test} & 81.2 & 81.2 & 80.3 & 80.2 & 80.1 \\
                        \midrule
                        \multirow{2}{*}{\textbf{OOD-3}}
                        & {\tt mtnt-2019} & 71.4 & 70.0 & 70.9 & 72.8 & 70.6 \\
                        & {\tt kftt} & 54.7 & 48.9 & 54.5 & 56.0 & 48.8 \\
                        \midrule
                        & {\tt average} & 72.3 & 70.7 & 71.5 & 71.8 & 69.9 \\
    				\bottomrule
    			\end{tabular}
    		\end{adjustbox}
			\label{tab:detection-ablation-intermediate}
		\end{center}
	\end{minipage}
\end{table} 

\subsection{Mismatched Pretrained Encoders}

This section investigates if it is possible to use alternative mismatched encoders as the backbone for a proxy system when predicting sequence-level attributes for the T5 model. We, therefore, investigate replacing the T5 encoder with RoBERTa \cite{roberta}, XLM-RoBERTa \cite{xlm-roberta} or the lightweight ALBERT \cite{albert}. See Table \ref{tab:mismatched-parameters} for information about the model size and inference time.

\begin{table}[h!]
	\centering{}
	\begin{minipage}[t]{1.0\textwidth}%
		\begin{center}
                \caption{Parameter counts and inference time of models on {\tt iwslt-2017}.}
			\def\arraystretch{1.20}
			\begin{adjustbox}{center}
    			\begin{tabular}{r|cccc|c}
    				\toprule
                        {\tt Layers} & {\tt Embeddings} & {\tt Encoder} & {\tt Head} & {\tt Total} & {\tt Inference Time} \\
                        \midrule
                        {\tt T5 Large Encoder} & 32.9M & 334.9M & 20.9M & 355.9M & 17.9s \\
                        \midrule
                        {\tt RoBERTa Base} & 39.0M & 124.1M & 11.8M & 135.9M & 4.3s \\
                        {\tt RoBERTa Large} & 52.0M & 354.3M & 20.9M & 375.3M & 17.5s \\
                        \midrule
                        {\tt XLM-RoBERTa Base} & 192.4M & 277.5M & 11.8M & 289.3M & 4.5s \\
                        {\tt XLM-RoBERTa Large} & 256.5M & 558.8M & 20.9M & 579.8M & 19.2s \\
                        \midrule
                        {\tt ALBERT Base} & 3.9M & 11.1M & 11.8M & 22.9M & 4.8s \\
                        {\tt ALBERT Large} & 3.9M & 16.6M & 20.9M & 37.6M & 19.4s \\
    				\bottomrule
    			\end{tabular}
    		\end{adjustbox}
			\label{tab:mismatched-parameters}
		\end{center}
	\end{minipage}
\end{table} 

The detection performance of alternative backbones such as base RoBERTa and base XLM-RoBERTa are slightly worse but with significantly lower inference times. The large RoBERTa and XLM-RoBERTa are approximately as fast as the T5 Encoder-based proxy but only the latter achieves similar detection performance. The lightweight ALBERT pretrained backbone significantly suffers at this task.

\begin{table}[h!]
	\centering{}
	\begin{minipage}[t]{1.0\textwidth}%
		\begin{center}
                \caption{Detection performance of NAPs using MI $\mathcal{I}$.}
                \def\arraystretch{1.20}
			\begin{adjustbox}{center}
    			\begin{tabular}{ll|c|cccccc}
    				\toprule
                        \multirow{2}{*}{\textbf{Split}} & \multirow{2}{*}{\textbf{Dataset}} & {\tt T5 Encoder} & \multicolumn{2}{c}{\tt RoBERTa} & \multicolumn{2}{c}{\tt XLM-RoBERTa} & \multicolumn{2}{c}{\tt ALBERT} \\
                        & & {\tt Large} & {\tt Base} & {\tt Large} & {\tt Base} & {\tt Large} & {\tt Base} & {\tt Large} \\
                        \midrule
                        \multirow{2}{*}{\textbf{OOD-1}}
                        & {\tt newstest-19} & \textbf{70.5} & 64.3 & 62.6 & 68.8 & 69.3 & 60.8 & 63.2 \\
                        & {\tt newstest-20} & \textbf{78.1} & 72.0 & 69.1 & 76.8 & 77.4 & 67.9 & 68.0 \\
                        \midrule
                        \multirow{2}{*}{\textbf{OOD-2}}
                        & {\tt khresmoi-dev} & 77.9 & 78.7 & 77.2 & 69.2 & \textbf{80.0} & 73.2 & 71.0 \\
                        & {\tt khresmoi-test} & 81.2 & 81.9 & 78.0 & 72.1 & \textbf{83.0} & 75.8 & 74.2 \\
                        \midrule
                        \multirow{2}{*}{\textbf{OOD-3}}
                        & {\tt mtnt-2019} & \textbf{71.4} & 61.6 & 62.1 & 61.7 & 61.6 & 63.5 & 68.3 \\
                        & {\tt kftt} & 54.7 & 61.7 & 62.1 & \textbf{62.6} & 62.3 & 51.4 & 43.0 \\
                        \midrule
                        & {\tt average} & \textbf{72.3} & 70.1 & 68.5 & 68.6 & \textbf{72.3} & 65.4 & 64.6 \\
    				\bottomrule
    			\end{tabular}
    		\end{adjustbox}
			\label{tab:detection-ablation-roberta}
		\end{center}
	\end{minipage}
\end{table} 

\subsection{Decorrelating Epistemic and Aleatoric Uncertainty}

Epistemic and aleatoric uncertainties are of different natures. The former is a measure of the lack of knowledge in our model parameters and model choice under the given dataset. As the dataset increases the epistemic uncertainty should decrease. The latter is an intrinsic measure of uncertainty in the data itself which might be caused by noisy data collection methods or labelling errors. Therefore, we propose a new loss function in which we aim to maximise the correlation between the proxy outputs $\{\hat{s}_i\}_i$ and teacher sequence-level epistemic scores $\{{s}_{ei}\}_i$ whilst also decorrelating its outputs from teacher sequence-level aleatoric scores $\{{s}_{ai}\}_i$:
\begin{align}
    \mathcal{L}_{\tt ep-al} = \mathcal{L}_{\tt scc}\Big(\{\hat{s}_i\}, \{{s}_{ei}\}\Big) - \alpha \Big\lvert{\mathcal{L}_{\tt scc}\Big(\{\hat{s}_i\}, \{{s}_{ai}\}\Big)}\Big\rvert
\end{align}
where $\alpha$ controls the level of decorrelation. Table \ref{tab:detection-ablation-mi-du} shows that by using this style of loss function, the proxy can be made to perform significantly better. The base model $\alpha = 0.0$ already outperforms a deep ensemble at detection, and furthermore, setting $\alpha = 1.0$ shows even better overall performance.

\begin{table}[h!]
	\centering{}
	\begin{minipage}[t]{1.0\textwidth}%
		\begin{center}
                \caption{Detection performance of NAPs using MI $\mathcal{I}$.}
                \def\arraystretch{1.20}
			\begin{adjustbox}{center}
    			\begin{tabular}{ll|cccc}
    				\toprule
                        \multirow{2}{*}{\textbf{Split}} & \multirow{2}{*}{\textbf{Dataset}} & \multicolumn{4}{c}{\textbf{NAP Large}} \\
                        & & $\alpha = 0.0$ & $0.5$ & $1.0$ & $2.0$ \\
                        \midrule
                        \multirow{2}{*}{\textbf{OOD-1}}
                        & {\tt newstest-19} & 70.5 & \textbf{76.1} & 76.0 & 75.3 \\
                        & {\tt newstest-20} & 78.1 & 85.9 & \textbf{86.3} & 84.0 \\
                        \midrule
                        \multirow{2}{*}{\textbf{OOD-2}}
                        & {\tt khresmoi-dev} & 77.9 & 86.1 & \textbf{88.0} & 83.5 \\
                        & {\tt khresmoi-test} & 81.2 & 86.8 & \textbf{87.7} & 83.3 \\
                        \midrule
                        \multirow{2}{*}{\textbf{OOD-3}}
                        & {\tt mtnt-2019} & \textbf{71.4} & 61.7 & 57.3 & 51.1 \\
                        & {\tt kftt} & 54.7 & 70.2 & 76.5 & \textbf{77.9} \\
                        \midrule
                        & {\tt average} & 72.3 & 77.8 & \textbf{78.6} & 75.9 \\
    				\bottomrule
    			\end{tabular}
    		\end{adjustbox}
			\label{tab:detection-ablation-mi-du}
		\end{center}
	\end{minipage}
\end{table}


\end{document}